- Source code and dataset available at: https://github.com/Zardoua-Yassir/A_fast_horizon_detector_and_a_new_annotated_dataset_for_maritime_video_processing

# A Fast Horizon Detector and a New Annotated Dataset for Maritime Video Processing

Yassir Zardoua [1,*], Mohammed Boulaala [2], Mhamed El Mrabet [2], and Abdelali Astito [1]

[1] Smart Systems and Emerging Technologies, Faculty of Science and Technologies of Tangier (FSTT), Abdelmalek Essaadi University, Tetouan, Morocco

[2] Industrial Systems Engineering and Energy Conversion Team, Faculty of Science and Technologies of Tangier (FSTT), Abdelmalek Essaadi University, Tetouan, Morocco

Email: yassirzardoua@gmail.com (Y.Z.); m.boulaala@gmail.com (M.B.); m.elmrabet@gmail.com (M.E.M.); abdelali_astito@yahoo.com (A.A.)

*Corresponding author

***Abstract*—Accurate and fast sea horizon detection is vital for tasks in autonomous navigation and maritime security, such as video stabilization, target region reduction, precise tracking, and obstacle avoidance. This paper introduces a novel sea horizon detector from RGB videos, focusing on rapid and effective sea noise suppression while preserving weak horizon edges. Line fitting methods are subsequently employed on filtered edges for horizon detection. We address the filtering problem by extracting line segments with a very low edge threshold, ensuring the detection of line segments even in low-contrast horizon conditions. We show that horizon line segments have simple and relevant properties in RGB images, which we exploit to suppress noisy segments. Then we use the surviving segments to construct a filtered edge map and infer the horizon from the filtered edges. We propose a careful incorporation of temporal in- formation for horizon inference and experimentally show its effectiveness. We address the computational constraint by providing a vectorized implementation for efficient CPU execution, and leveraging image downsizing with minimal loss of accuracy on the original size. Moreover, we contribute a public horizon line dataset to enrich existing data resources. After extensive tests, we report the following major findings: 1) thanks to its filter, our algorithm accurately detects horizon lines with low or weak edge response, 2) the vectorized filter takes no more than 1.71% of the overall computations, while most of the computations are taken by the Line Segment Detection (LSD) algorithm we integrated into our pipeline, 3) our strategy of incorporating the temporal information avoids outlier detections, mitigates the effect of strong noisy lines, and exhibits high robustness when using incorrect detections as a temporal reference. Our algorithm's performance is rigorously evaluated against state-of-the-art methods, and its core components are validated through ablation experiments.***

Manuscript received January 26, 2024; revised April 12, 2024; accepted May 13, 2024; published October 18, 2024.

## I. Introduction

Video surveillance has become a vital tool for maintaining security and safety in various domains. Traditional systems rely primarily on manual human observation, typically from Close-Circuit Television (CCTV). The recent progress in computer vision and deep learning revolutionized this field, allowing automated surveillance with remarkable performance [1, 2]. Automated crowd counting enables flow monitoring and deploying security services [3]. During the recent pandemic, visual social distancing has been introduced as a new computer vision problem to mitigate and manage the spread of infectious diseases [4]. Other surveillance tasks include the analysis of video captured from different non-overlapping cameras, such as person reidentification, aiding the identification of suspects or lost children [5].

Within this realm, maritime target detection and tracking systems significantly aid in maritime security. The literature decomposes such systems into multiple algorithms [6, 7], such as maritime background modeling [8], video stabilization [9], and maritime horizon detection [10]. Our paper contributes to the maritime video surveillance system by developing and testing a fast and robust horizon detection algorithm. The maritime horizon, also referred to as the sea-sky line in some cases, is defined as the boundary line that separates the sea region from the region immediately above it (Fig. 1) [6, 11, 12]. We should note that in maritime images depicting mountains or coastlines, as illustrated in Fig. 1(b), the horizon according to the maritime video processing literature is the line separating the mountain from the sea [6, 7, 13–16]. This maritime horizon is semantically distinct from the wild horizon literature, which addresses the non-linear boundary separating the mountain from the sky [17, 18].

Our paper focuses on detecting the maritime horizon line in RGB images, which, as highlighted in [7], offer richer daytime information at a lower cost compared to infrared images. This sea horizon plays a pivotal role in intelligent and autonomous sea navigation [6, 7]. Previous studies underscored the application benefits of correctly detecting the sea horizon. These applications include digital video stabilization [9, 19, 20], improved target detection and tracking [13, 21], reduced search region [22–24], and obstacle avoidance [25–27]. We note that these studies include commercial surveillance systems like Automatic Sea Vision (ASV®) [28], which integrate the sea horizon detector. Survey [12] reports a strong trade-off between robustness and computational load. Therefore, further research in this specific field is necessary.

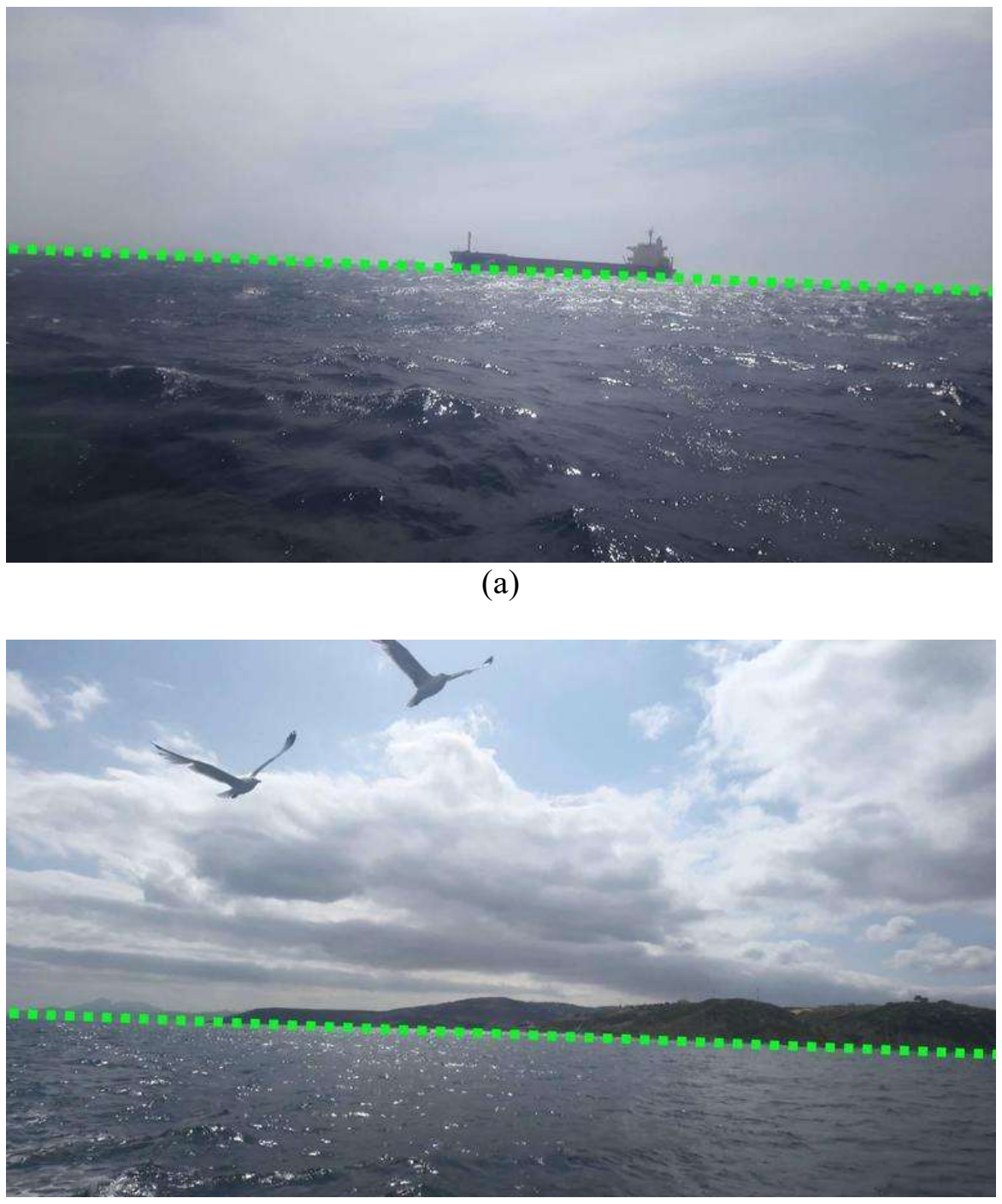

(a)

(b)

Fig. 1. The horizon line separates (a) the sea from the sky and (b) the sea from the coast.

Previous and recent research consistently underscores the essential role of edge information in accurately detecting the horizon line [6, 7, 14, 29]. The analysis in Section II and survey [12] suggest that the current literature would greatly benefit from a horizon detector that meets two critical criteria: (1) high robustness against sea clutter while preserving weak horizon edges and (2) efficient CPU execution with minimal computations [1]. Accordingly, we categorize our contributions into algorithmic and dataset contributions. We summarize our algorithmic contributions in the following four points:

(1) We introduce a novel filter that generates a filtered edge map, preserving very weak horizon edges while suppressing various forms of sea clutter. This filtered edge map is then used for horizon line inference;
(2) In the line inference stage, we propose an Outlier Handler Model (OHM) that effectively incorporates temporal information to avoid outlier detections;
(3) We enhance the computational efficiency by vectorizing the filtering processes;
(4) We effectively leverage image downsizing for faster computations by accurately projecting the detected horizon from the downsized image to the original size.

Our dataset contribution involves:

(1) the public sharing of 18 video clips, totaling over 6,000 frames, which we meticulously annotated for the maritime horizon line;
(2) introducing multiple new scenarios to enhance and complement the most extensive video dataset currently available, the Singapore Maritime Dataset (SMD), initially introduced in [6].

## II. Literature Review

In this section, we succinctly discuss an extensive set of horizon detection algorithms. We recommend that readers refer to surveys [6, 7, 12] for an in-depth literature analysis. Gershikov *et al.* [30] detect one horizon point per column as the pixel with the maximum vertical edge response. The final horizon is fit on these points using the least-squares technique, which is known for its high sensitivity to outliers. Better algorithms project edge maps to another space favoring the appearance of the horizon as a peak. The first attempt in this context was conducted by Bao *et al.* [31]. They detected image edges using LoG (Laplacian of Gaussian) and extracted the horizon using the global maximum of the transformed space. The horizon may not correspond to the global maximum due to image noises. Therefore, Zhang *et al.* [32] enhanced the work in [31] by analyzing three geometric features of local peaks of the transformed space. These features assume that noisy edges have high scatter, which is unsuitable for line-up sea clutter. Similar to Ref. [31], Schwendeman and Thomson [20] detected the horizon as the global peak of the Hough space and suggested a contrast-based quality metric to remove faulty global peaks. Schwendeman *et al.* [20] report that their method is susceptible to lighting changes, one of the main factors degrading horizon edges. Similar to Ref. [20], Lie *et al.* [17] employed a Canny edge detector with an automatic threshold to extract the weak horizon edges while minimizing the noise. Subsequently, they identified the horizon using a dynamic programming technique to infer the curved horizon from the extracted edges. While they reported good results on low-contrast horizon of non-maritime images, the effectiveness of this methodology in our case has yet to be evaluated.

Many researchers have observed that directly inferring the horizon from extracted edges often leads to incorrect detections, primarily due to high texture in RGB images [6, 12]. As a result, they have incorporated various

[1] The reader can find a discussion on the importance of criteria (2) in survey [12]

edge filtering techniques to enhance clutter suppression while preserving horizon edges. In this context, Shen *et al.* [10, 33] proposed an adaptive size of the Gaussian kernel to avoid excessive suppression of horizon edges. The size is established based on the image resolution, specifically the number of rows. Regardless of the Gaussian kernel's window size, it consistently weakens the horizon edges due to the kernel's built-in smoothing effect, reducing high-frequency components in the image. The authors of [30, 34] have discovered that morphological erosion is more effective in preserving weak horizon edges compared to Gaussian filters. In response to this, Li *et al.* [15] recently introduced a more advanced morphological filter that calculates the reconstruction by erosion of the edge response. This filter operates under the assumption that maritime clutter in RGB images typically manifests as blobs, such as sun glints on the sea surface or small waves. This assumption allows the filter to effectively preserve horizon edges, which do not appear as blobs. However, the iterative nature of the erosion-based reconstruction technique, even when applied to downsized images, impacts its processing speed. Numerous researchers have conducted extensive investigations into the utility of median filters. These filters have garnered attention for their superior performance compared to Gaussian filters, primarily due to their capacity to preserve intensity gradients without excessive smoothing. Median filters, with odd-sized pixel windows, prevent the creation of intermediate intensities that could weaken the transition across the horizon edge. They achieve this by ranking pixel intensities and selecting the central value, preserving a sharp transition across the horizon line [35].

To increase the suppression effect, it is better to use multiple median windows of increasing size instead of one large median filter [36]. This is known as the multi-scale median filter, an approach that has produced some of the best results in the maritime literature [29, 31, 37–39]. In alignment with this approach, Prasad *et al.* [38] applied five median scales and transformed the result into a Hough space where the voting rule is modified to favor longer edges. Such a process is resource-greedy and takes tens of seconds per frame. A similar and faster alternative applies ten median scales [37] and detects horizon candidates from each scale using the vertical edge response and the standard Hough transform. The final candidate is selected using a goodness score based on the strength and collinearity of its edges. Instead of detecting candidate lines on each median scale, Jeong *et al.* [39] significantly reduced the computational load of the multi-scale median filter by processing only one weighted edge map, which they computed by averaging the edges of each median scale. They conducted experiments that demonstrated this approach's significantly improved accuracy and remarkable speed enhancement, being 90 times faster than the methods presented in [37] and [38]. Overall, multi-scale median filtering [39] and the sophisticated morphological filter [15] reported the best ability to preserve weak horizon edges.

Dong *et al.* [14] proposed a novel filter that effectively removes sea noise while detecting horizon lines with weak edges. They addressed computational speed concerns by extracting a Region of Interest (ROI) encompassing the horizon. Line segments were subsequently extracted from the ROI using an adaptive local threshold. The authors of this filter conducted experiments to demonstrate that the distribution of gradient angles along horizon line segments is significantly higher than that of the noise. Therefore, they stretched the line segments by a factor of 0.5, filtered them based on the stability of their gradient angles, and applied a RANSAC algorithm to obtain the horizon line. The same Ref. [14] justifies the 0.5 stretching by explaining that the horizon line segments will further span along the horizon, thereby enhancing their gradient angle stability while reducing the stability for noisy segments. This filtering approach effectively preserved weak horizon edges and outperformed median-scale filters [37, 39]. However, its applicability to our case is questionable because this algorithm was designed based on the unique properties of infrared imagery, which significantly differ from RGB images, as detailed in [6, 7]. For example, studies in [6, 14, 40] point out that the infrared texture on the sea surface is poorer compared to RGB images. This characteristic is advantageous for streamlining the filtering process since higher texture levels tend to introduce more noisy segments.

Another category of algorithms incorporates regional properties such as pixel intensities and texture, which are usually fused with edge-based features to improve the accuracy and computational load. For instance, Jeong *et al.* [39] process only a region of interest extracted through analysis of color distribution difference of multiple sub-images. Other methods exploit the color properties of the sky to detect it and assume that the horizon is the linear boundary right below it [41, 42]. Such assumption easily breaks on images depicting coastal regions (Fig. 1(b)), where the horizon is the boundary separating the coast from the sea rather than the sky class. Ettinger *et al.* [43] and Fefilatyev *et al.* [21] consider the horizon as the line maximizing the intra-class variance of the two regions split by that line, which requires expensive computations even on low-resolution images. Liang and Liang [16] significantly enhanced the speed and accuracy of this approach by processing color and texture information in two small patches sliding along the considered Hough candidate.

Simple assumptions about the color and texture of maritime semantic classes lead to failures even for machine learning-based algorithms, such as Support Vector Machine (SVM), decision trees [44], and Gaussian Mixture Model (GMM) [22, 45]. For instance, Kristan *et al.* [45] train a GMM where the feature vector of each class (sea, coast, and sky) is supposed to have a Gaussian distribution, which is unsuitable due to the non-homogeneity of image classes (details in [12]). The machine learning paradigm has been investigated for use on non-maritime images as well. In these cases, the horizon is represented as a non-linear boundary separating ground and sky pixels. Some learning-based methods, such as those described in [46, 47], process only specific parts of the image. These methods focus on image patches centered at pixels identified by the Canny edge detection algorithm. Classifying all Canny edge pixels without applying a proper filter is time consuming, especially if we lower the edge detection threshold for extracting weak horizon edges. The edge classification approach was quickly adopted for maritime images by Jeong *et al.* [39], who demonstrated that using a convolutional neural network outperformed traditional classifiers and handcrafted features [46, 47]. To reduce computational load, they processed the input image with the multi-scale median filter from [39], employing two additional scales to limit the number of candidate edge pixels and speed-up the classification process. While the study in [29] reports excellent CNN performance on a wide range of maritime images, we found that the excessive use of median filtering scales suppresses not only weak horizon edges but also those with higher edge magnitudes. Therefore, the horizon detector in [29] is limited to horizon lines with prominent edges.

## III. Materials and Methods

### *A. The Proposed Dataset*

An ideal sea horizon dataset should contain carefully annotated video clips depicting a comprehensive set of maritime conditions. The horizon annotation is necessary for the quantitative evaluation of a given algorithm. The annotation files provide GT values of the horizon position $Y$ and tilt $\phi$, as shown in Fig. 2.

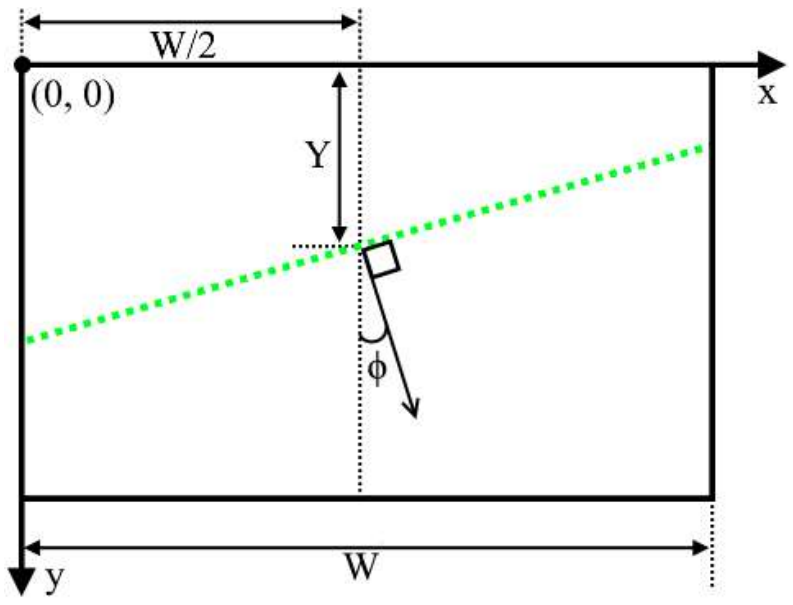

Fig. 2. position $Y$ and tilt $\phi$ of the horizon line (dotted green line); W represents the image width.

There are several maritime datasets such as [21, 45, 48]. However, they lack high-resolution images [21, 45], horizon line annotations [45], diversity of conditions [21], temporal information (i.e., consecutive video frames) [45, 48], or low altitude images. The last case, i.e., low altitude images, is necessary as it reflects the conditions faced by surface marine vehicles [48]. As far as we know, the SMD [6] is the largest and most diverse public dataset without the previous drawbacks. Student volunteers [2] annotated the SMD, which was collected using two setups: the onshore and onboard setup[3].

The SMD lacks important maritime scenes, which have the potential to break assumptions of state-of-the-art algorithms. To address this gap, we introduce the Tangier Maritime Dataset (TMD), which comprises 18 high-resolution, visible-range videos captured onboard a surface vessel navigating the Mediterranean Sea near Tangier city, Morocco. We carefully annotated the TMD using a custom software that we developed in Python and Tkinter [49] (Fig. 3).

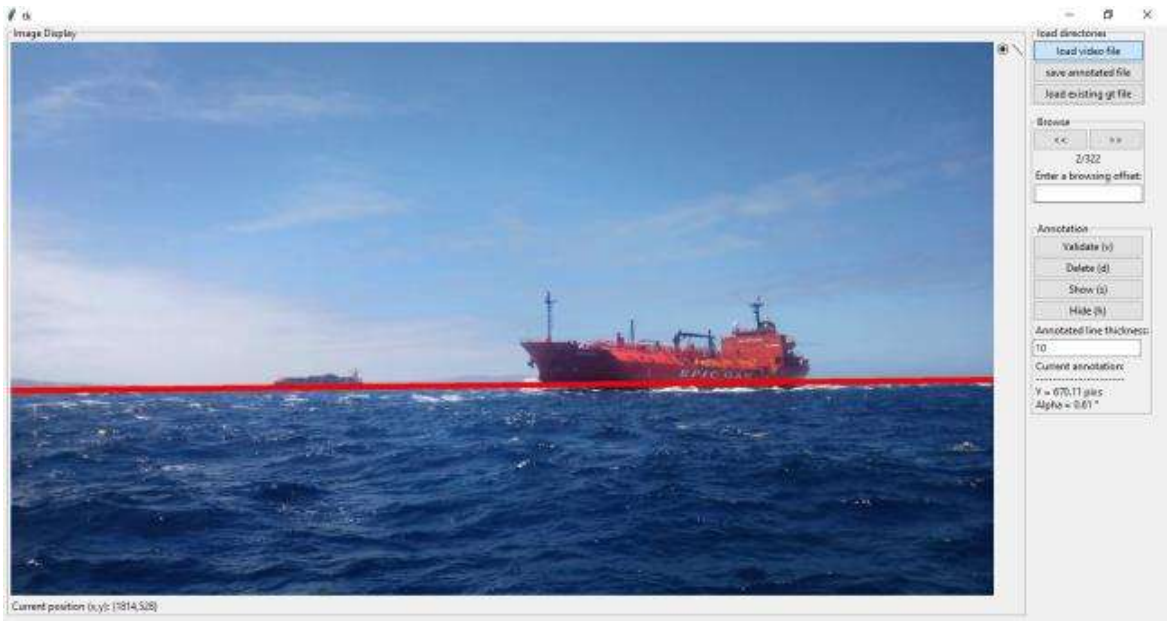

Fig. 3. A snap of the GUI of our annotation software after annotating the horizon line (in red).

We verified the accuracy of each frame annotation. To suit both Python and Matlab users, we share the horizon annotations as .npy and .mat files with a description file (see Description of the dataset.pdf in the provided link). We highlight in Table I major maritime challenges existing in the TMD and SMD; our dataset contains nine maritime scenarios that do not exist in the SMD. Fig. 4 shows one sample frame from each video clip of the proposed TMD. Table II shows video properties and statistical details of the datasets we used in Section IV.

TABLE I. Diversity Comparison Between the Tangier Maritime Dataset with the Singapore Maritime Dataset. S, T, and B Mean that the Condition Exists in, the SMD Only, the TMD Only, and Both SMD and TMD, Respectively

| Condition Category | Condition Description | Dataset Presence |
|---|---|---|
| Sea variables | 1. Small waves | B |
| | 2. Scattered sun glints | B |
| | 3. Grouped sun glints | T |
| | 4. Smooth surface | T |
| | 5. Wakes | B |
| Sky variables | 6. Clear | B |
| | 7. Small clouds | B |
| | 8. Linear-edged clouds | T |
| | 9. Smooth overcast | B |
| | 10. Textured overcast | T |
| | 11. Visible Sun | T |
| Horizon Visibility | 12. Medium contrast | B |
| | 13. Weak horizon edges | T |
| | 14. Partial ship occlusion | B |
| Classes above the horizon | 15. Sky | B |
| | 16. Coast | T |
| | 17. Sky and coast | T |
| Time of the day | 18. Sunrise and sunset | T |
| | 19. Daytime | B |
| | 20. After sunset | S |

[2] We had to correct several incorrect annotations of the SMD before using it in the experiments.

[3] Onshore means that the camera is mounted on a stationary shore platform, whereas onboard means that the camera is mounted on a offshore non-stationary platform (e.g., buoy or a ship).

TABLE II. VIDEO PROPERTIES AND STATISTICAL DETAILS DATASETS

| | SMD | | TMD (ours) |
|---|---|---|---|
| | **Onboard** | **Onshore** | |
| Number of video files | 11 | 37 | 18 |
| Total frames | 2813 | 17222 | 6090 |
| Frames per video | ∈ [255, 299] | ∈ [206, 995] | ∈ [273,405] |
| Frame size (height × width) | 1920 × 1080 | 1920 × 1080 | 1920 × 1080 |
| FPS | 30 | 30 | 30 |
| Channels | RGB | RGB | RGB |

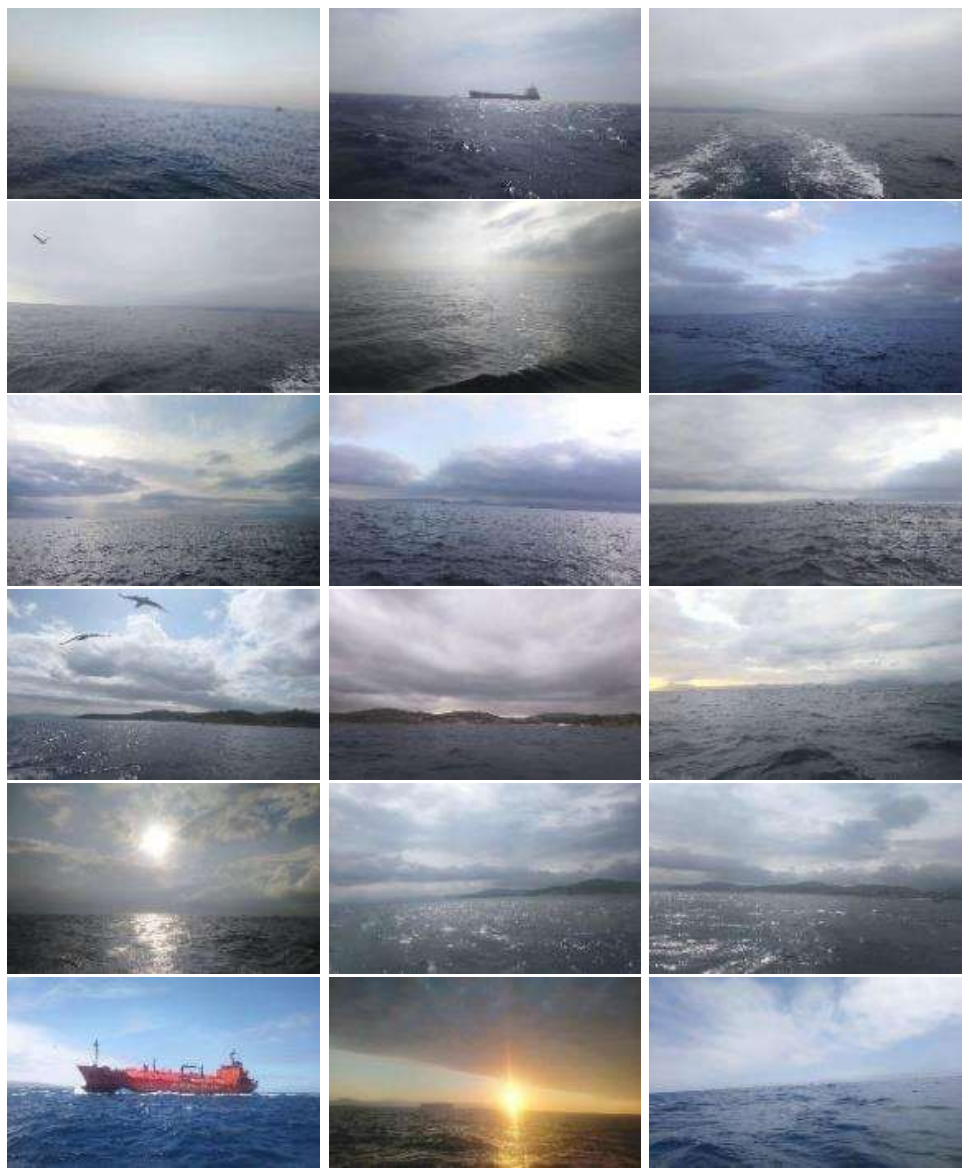

Fig. 4. One sample frame from each video clip of the TMD.

## B. Algorithm Overview

We will provide an overview of the proposed algorithm in this section before detailing its stages in Sections III.C, III.D, III.E, and III.F. We show the overall algorithm pipeline in Fig. 5. We note that the main ideas we employed in Stages 2 and 3 (Fig. 5) were introduced in the recent survey [12] as a promising research direction. The study in [12] motivated these ideas after examining the horizon line segment properties in 85 RGB images with challenging sea clutter. Stage 1 extracts the red channel[4] of the input image $I_{rgb}(x, y)$ and applies a Line Segment Detector (LSD) algorithm to detect the set of line segments $S^a$. The key process is to select from $S^a$ Candidate Horizon Segments (CHSs), which are more likely to correspond to the horizon line. We denote CHSs by set $S^f$. We rely on two filtering stages in establishing set $S^f$ : the Length-Slope Filter (LSF) and the Region of Interest Filter (ROIF) (Fig. 5). The LSF selects from $S^a$ two sets of segments, $S^c$ and $S^d$, which contain the longest segments with a slope less than a threshold $\alpha_{th}$. We directly consider segments of set $S^c$ as CHS, i.e., $S^c \subset S^f$, because they are all longer than segments of set $S^d$. The ROIF establishes one tight Region of Interest (ROI) that encompasses each segment of set $S^c$. Subsequently, a segment of set $S^d$ is selected as an additional CHS only if encompassed by at least one of the defined ROIs. We denote segments satisfying such a condition by $S^e$. We denote the set containing all CHSs by $S^f$, which is equal to $S^c \cup S^e$. Stage 4 establishes the filtered edge map $E(x, y)$ as pixels along each segment of set $S^f$. Finally, Stage 5 infers the horizon using line-fitting techniques and temporal information.

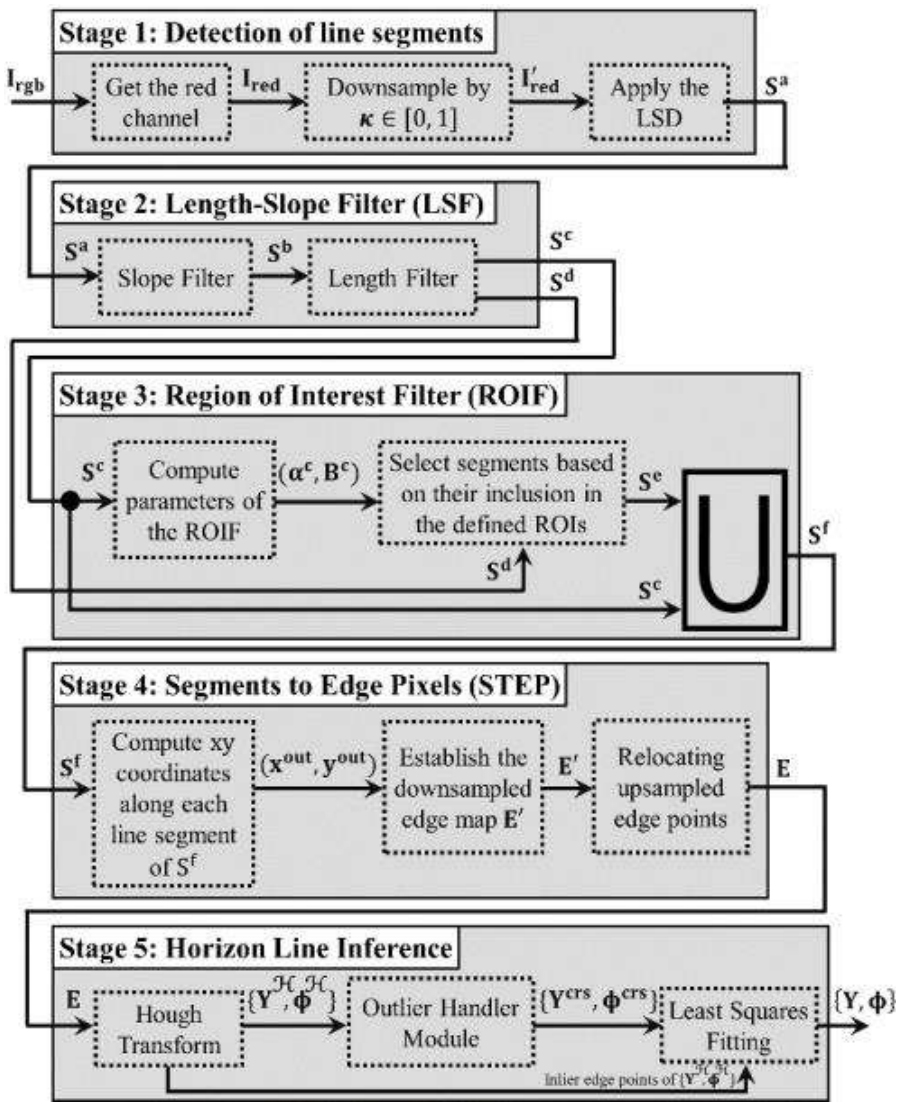

Fig. 5. Pipeline of our algorithm.

## C. Detection of Line Segments

Our method detects and evaluates image line segments instead of whole lines to create a filtered edge map $E(x, y)$ with relevant horizon edges. The use of line segments provides numerous benefits. First, we gain more robustness against partial occlusions of the horizon (e.g., by ships). Such occlusion significantly changes horizon properties in occluded regions, making the latest line detectors ineffective against simple occlusions, as highlighted in [12]. Second, line segments are inherently efficient scale-invariant descriptors [50] and can be shared with other components of the system for efficient video processing. Third, as we will show in Section III.D and Section III.E, horizon line segments exhibit simple and relevant properties on a wide range of cluttered RGB images. Fourth, detecting line segments is an old problem, and the current literature offers several fast and robust algorithms with sub-pixel accuracy [51, 52].

In this paper, we detect line segments using the fast Line Segment Detector (LSD) developed by Gioi *et al.* [51], which combines the advantages of its antecedents while significantly limiting their inconveniences. We exploit its following properties to create a robust and fast horizon edge filter: not misled by faulty segments induced by highly textured regions thanks to the gradient angle; a significant decrease of the false positives and negatives related to fixed detection thresholds; adopts a novel criterion selecting both long and short segments; satisfying linear-time execution. As shown in Fig. 5, we extract the red channel and down-sample it by $\kappa \in [0, 1]$. Applying the LSD on the down-sampled red channel $I_{red}$ reduces the computational time and mitigates the negative effect of high-frequency components. Throughout this paper, we denote an unknown set of segments $u$ by $S^u$, its number of elements by $N^u$, and the $v$-th segment of set $S^u$ by $S^{u,v}$. Thus, we can express a given set $u$ as $S^u = \{S^{u,v}\}_{v=1:N^u}$. Given this notation, we refer to segments output by the LSD by $S^a = \{S^{a,i}\}_{i=1:N^a}$. Fig. 6(b) shows that despite the weak edge response of the horizon (Fig. 6(a)), set $S^a$ contains enough horizon segments to allow its detection. We explain this result by the relative stability of gradient angles on the horizon, which allows the LSD [51] to grow longer segments on the horizon even when it is blurred. We favor such a possibility by setting a small gradient magnitude threshold for the LSD[5].

[4] We applied the LSD on 9 color channels on a wide range of sea images, visually analyzed the results, and concluded that the red channel produced the best results.

### *D. Length-Slope Filter (LSF) and Description and Motivation*

The LSF (stage 2 in Fig. 5) filters segments of set Sa using their slope and length. In our case, the EO sensor is mounted on terrestrial moving platforms whose deviation is much smaller than aerial-based platforms. Therefore, we suppress too tilted line segments, which may lead to detecting false sea horizons when they are remarkably long. Several factors can induce too-tilted segments, such as the vehicle's wakes depicted by the rear camera or the long vertical edges (e.g., antennas) of nearby ships. Therefore, the LSF first filters out segments of set Sa according to the slope condition of Eq. (1).

$$S^{a,i} \in S^b \text{ if } \left|\alpha^{a,i}\right| \leq \alpha_{th} \tag{1}$$

where $|\,.\,|$ is the absolute value, $\alpha^{u,v}$ denotes the slope of segment $S^{u,v}$, $\alpha_{th}$ is a scalar threshold, and $S^b$ is the set of segments satisfying the slope condition. Fixing $\alpha_{th}$ to a minimal value (e.g., $\alpha_{th} = 0.009$) can be tempting. However, doing so may eliminate valuable horizon segments. This is because cameras on small sea platforms, such as buoys and USVs, often experience significant deviations due to sea winds and waves, as noted in previous studies [6, 21, 45]. We further filter segments of set $S^b$ by removing the relatively shorter segments. Concretely, segment $S^{b,j}$ will survive the filter only if it is among the longest $N^c$ segments of set $S^b$. We formally express this condition in Eq. (2).

$$S^{b,j} \in S^c \text{ if } L^{b,j} \geq L_{srt}^{b,N^c} \tag{2}$$

where $S^c$ is the set containing survived segments, $L^{u,v}$ is the length of the $v$-th segment $S^{u,v}$, $N^c$ is a scalar threshold specifying the number of the longest segments to select[6], and $L_{srt}^{u,v}$ is the length (in pixels) of the $v$-th-longest segment[7] of set $S^u$. The underlying assumption of the filtering condition of Eq. (2) is that pixels of the horizon line will ideally have the same gradient orientation value, which allows the LSD to group horizon pixels into the lengthiest segments. In contrast, pixels of the most common sea clutter (e.g., sea waves, sun glints, wakes) will correspond to scattered gradient orientations. In light of this, Fig. 6(b) shows that despite a large number of line segments on the sea surface, most are shorter than the few segments detected on the horizon. Hence, we perform no further filtering on segments of set $S^c$ and consider them as horizon segments, i.e., $S^c \subset S^f$ (see the output of Stage 3 in Fig. 5). Fig. 6(c) shows segments of set $S^c$ for $N^c = 15$. We explain in Section E that settling for set $S^c$ as the only horizon segments could lead to missing valuable horizon segments due to some factors affecting the stability gradient orientations of horizon pixels. Therefore, we select more segments by taking the next longest $N^d$ segments from set $S^b$ as expressed in Eq. (3).

$$S^{b,j} \in S^d \text{ if } L_{srt}^{b,N^c} > L^{b,j} \geq L_{srt}^{b,N^{d'}} \tag{3}$$

where $N^{d'} = N^c + N^d$ and $N^d$ is a scalar threshold that corresponds to the number of segments in set $S^d$ ($N^d \gg N^c$; e.g., $N^d = N^c \times 10$). Unlike set $S^c$, we do not directly consider segments in set $S^d$ as horizon segments because their length is relatively close to that of noisy segments, as shown in Fig. 6(d). The next stage captures additional horizon segments from set $S^d$ using information extracted from set $S^c$.

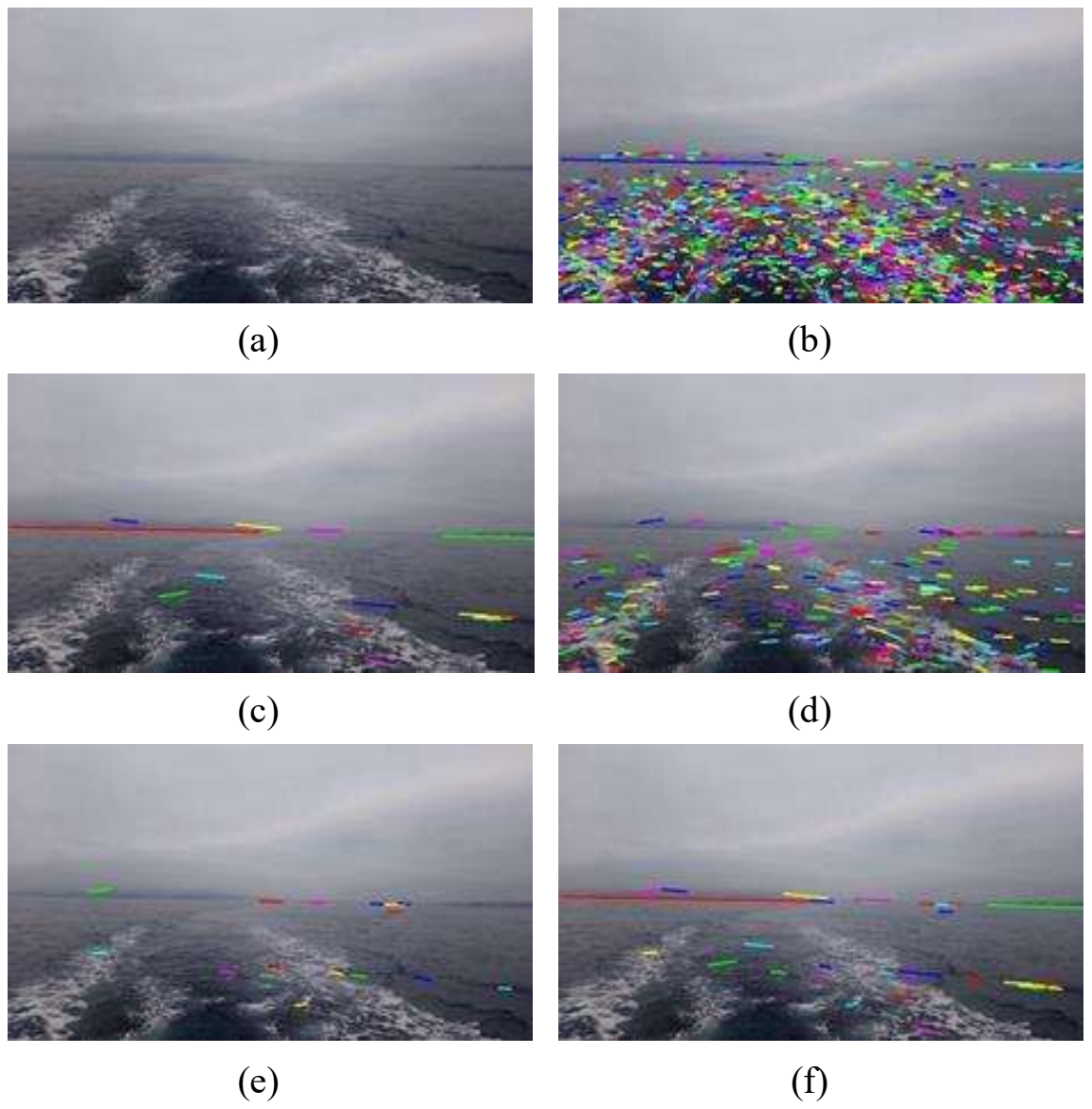

Fig. 6. Intermediate results of line segments filtering: (a) original image; (b) segments of set $S^a$; (c) segments of set $S^c$; (d) segments of set $S^d$; (e) segments of set $S^e$; (f) segments of set $S^f$.

[5] To be precise, we talk about the lowest hysteresis threshold.

[6] $N^c$ is also the number of elements (segments) in set $S^c$.

[7] For instance, $L_{srt}^{b,1}$ is the length value of the longest segment in set $S^b$.

### E. Vectorizing the Length-Slope Filter (LSF)

We denote Cartesian coordinates of the starting point and ending point of the $v$-th segment $S^{u,v}$ of set $S^u$ by $p_s^{u,v}: (x_s^{u,v}, y_s^{u,v})$ and $p_e^{u,v}: (x_e^{u,v}, y_e^{u,v})$, respectively. Thus, we denote $x_s^u$, $x_e^u$, $y_s^u$, and $y_e^u$ as vectors containing coordinates of segments of set $S^u$:

$$\begin{aligned} x_s^u &= \left[x_s^{u,1}, x_s^{u,2}, \dots, x_s^{u,N^u}\right]^T \in \mathbb{R}^{N^u \times 1} \\ x_e^u &= \left[x_e^{u,1}, x_e^{u,2}, \dots, x_e^{u,N^u}\right]^T \in \mathbb{R}^{N^u \times 1} \\ y_s^u &= \left[y_s^{u,1}, y_s^{u,2}, \dots, y_s^{u,N^u}\right]^T \in \mathbb{R}^{N^u \times 1} \\ y_e^u &= \left[y_e^{u,1}, y_e^{u,2}, \dots, y_e^{u,N^u}\right]^T \in \mathbb{R}^{N^u \times 1} \end{aligned} \tag{4}$$

There are two processes to vectorize in the LSF: the slope-based (Eq. (1)) and length-based (Eqs. (2) and (3)) filtering conditions. The former starts by computing vector $\alpha^a$ according to Eq. (5).

$$\begin{aligned} \alpha^a &= \left[\alpha^{a,1}, \alpha^{a,2}, \dots, \alpha^{a,N^a}\right]^T \in \mathbb{R}^{N^a \times 1} \\ &= (y_e^a - y_s^a) \oslash (x_e^a - x_s^a) \end{aligned} \tag{5}$$

where $\alpha^u$ is the vector corresponding to the slopes of segments in $S^u$ and $\oslash$ is the Hadamard (component-wise) division. All filtered sets of segments in this paper ($S^b$, $S^c, S^d, S^e$, and $S^f$) represent subsets of set $S^a$. We establish these filtered sets using vectorized indexing. Concretely and generally, we use the different filtering conditions (e.g., Eqs. (1) and (2)) to establish the vector $I_{u',u} \in \mathbb{R}^{N^{u'} \times 1}$. The elements of $I_{u',u}$ are used to index set $S^u$, which produces the filtered set $S^{u'}$. We will see in SectionIII.E.4 that vector $I_{u',u}$ also allows obtaining attributes of a given set $S^{u'}$ from homologous attributes of one of its supersets [8] $S^u$, avoiding thus redundant computations. Using discussed indexing, we get segments of set $S^b$ by establishing vector $I_{b,a} \in \mathbb{R}^{N^b \times 1}$, whose values are indices of slopes $\alpha^{a,i}$ (Eq. (5)) satisfying the condition expressed in Eq. (1). Eventually, we index set $S^a$ by vector $I_{b,a}$ to produce the desired set $S^b$. As set $S^c$ and $S^d$ are established based on the length of segments of set $S^b$, we first compute a length vector according to Eq. (6).

$$L^b = \left[L^{b,1}, L^{b,2}, \dots, L^{b,N^b}\right]^T \in \mathbb{R}^{N^b \times 1} = \left((x_e^b - x_s^b)^{\odot 2} - (y_e^b - y_s^b)^{\odot 2}\right)^{\odot \frac{1}{2}} \tag{6}$$

where $L^{u,v}$ is the length of the $v$-th segment $S^{u,v}$ and $\odot$ is the Hadamard (element-wise) power. Then, we establish vector $I_{srt}^{L^b} \in \mathbb{R}^{N^b \times 1}$ whose values are indices that would sort values of $L^b$ from highest to lowest. Thus, vector $I_{c,b}$, whose elements are indices of set $S^b$ satisfying Eq. (2), is established by slicing the first $N^c$ elements of vector $I_{srt}^{L^b} \in \mathbb{R}^{N^b \times 1}$. Similarly, we get vector $I_{d,b}$, whose elements are indices of set $S^b$ satisfying Eq. (3), by slicing $I_{srt}^{L^b}$ from the $(N^c + 1)$-th element to the $(N^c + 1 + N^d)$-th element. Eventually, we get set $S^c$ and $S^d$ by indexing set $S^b$ using vectors $I_{c,b}$ and $I_{d,b}$, respectively.

[8] For instance, we can get the slope vector $\alpha^{u'}$ by indexing $\alpha^u$ with vector $I_{u',u}$.

### F. Region of Interest Filter (ROIF)

#### 1) Motivation and assumtion

We mentioned in Section III.D that taking the longest $N^c$ segments from set $S^b$ could miss valuable line segments. This issue is fundamentally caused by certain types of noise, such as the Gaussian noise of low-quality sensors and the poor horizon contrast. Such noise disturbs the gradient orientation of horizon pixels, thus hindering the LSD from growing long segments on the horizon line. Fig. 6(d) shows that the low contrast condition led to growing short segments on the horizon. The same disturbed property, i.e., gradient orientation, provides the key to mitigating this issue; we assume that, unlike noisy segments, short horizon segments have a much higher collinearity with longer horizon segments. Fig. 7(a) depicts a synthetic example of this assumption. We show the result of using this assumption on the real image in Fig. 6(a). Fig. 6(e) shows that additional horizon segments, denoted as set $S^e$, are successfully captured from $S^d$. Selecting more horizon segments from $S^d$ often leads to capturing more noisy segments, as shown in Fig. 6(e). However, having more horizon segments allows more performance, as we will experimentally show in Section IV.F.

#### 2) The filtering condition: Initial formulation

We aim in this Section to formulate an initial expression of the filtering condition based on the collinearity assumption we mentioned in Section III.D.1 and illustrated in Fig. 7(a). To this end, we define one region of interest, denoted as $\mathcal{R}^{c,k}$, to encompass the k-th segment $S^{c,k}$. Subsequently, we consider that the $l$-th segment $S^{d,l}$ is an additional horizon segment only if both of its endpoints fall in at least one and the same region $\mathcal{R}^{c,k}$. Fig. 7(b) shows an example of the filtering regions corresponding to Fig. 7(a) in pink. Fig. 7(c) shows in yellow the segments of set $S^d$ satisfying the condition we just mentioned. We initially formulate this condition in Eq. (7).

$$\begin{aligned} &S^{d,l} \in S^e \text{ if } \exists\, k : \\ &\left(\left(p_s^{d,l} \in \mathcal{R}^{c,k}\right) \wedge \left(p_e^{d,l} \in \mathcal{R}^{c,k}\right)\right) = 1 \end{aligned} \tag{7}$$

where $S^e$ is the set grouping additional horizon segments selected from set $S^d$, $p_s^{d,l}$, and $p_e^{d,l}$ are the starting and ending point of segment $S^{d,l}$, respectively. $\in$ is the inclusion operator, $\wedge$ is the logical *and*, and 1 is the true Boolean.

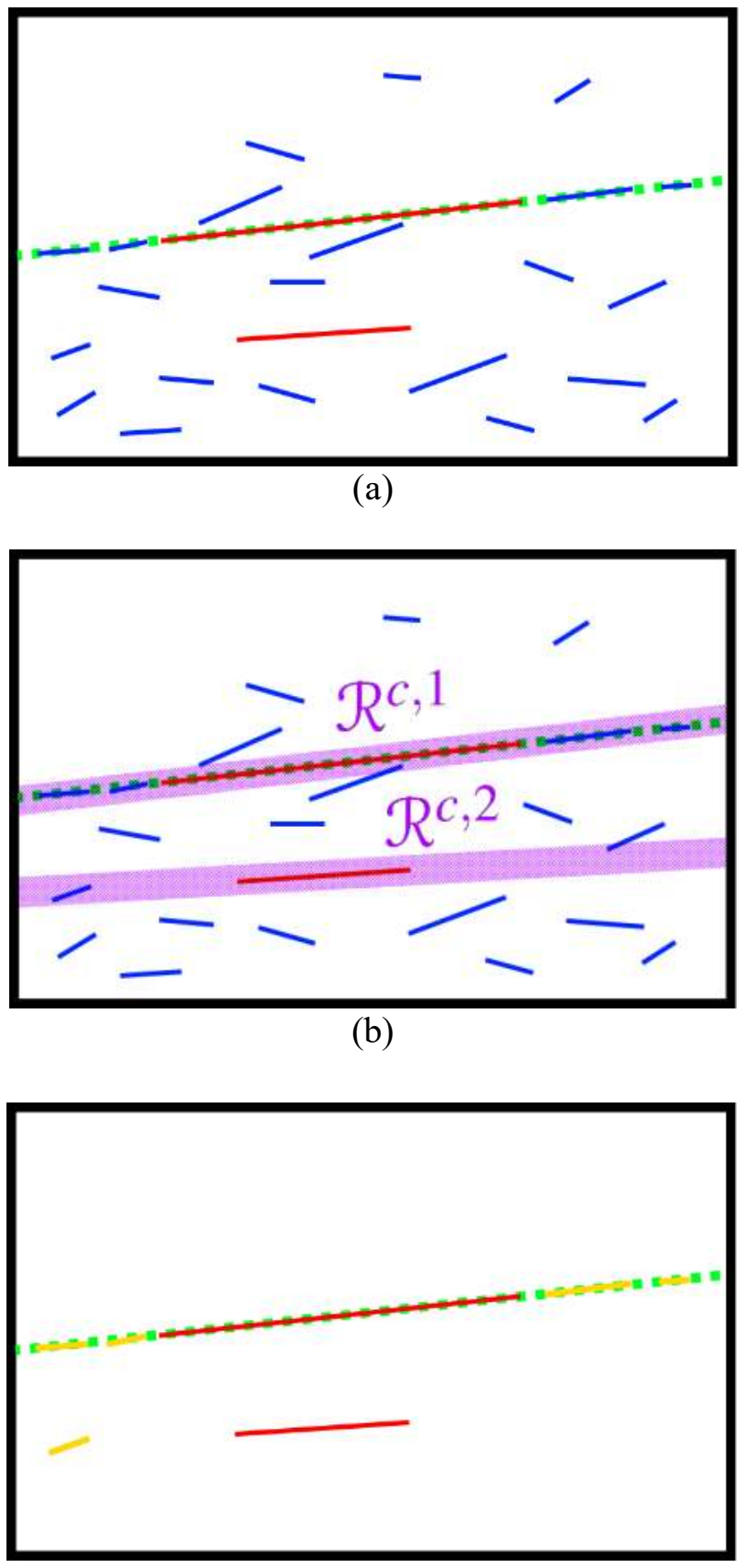

Fig. 7. A synthetic example showing the assumption made in the region of interest filter and its usage; the green dotted line is the horizon line. (a) segments of set $S^c$ (in red) and $S^d$ (in blue), (b) the filtering regions $\mathcal{R}^{c,1}$ and $\mathcal{R}^{c,2}$ (in pink) encompassing the two segments of set $S^c$, (c) segments of set $S^d$ that survived the two filtering regions (in yellow).

*3) Mathematical development of the filtering condition*

Implementing the filtering condition in Eq. (7) requires further mathematical development. This breaks down to verifying that a given endpoint of segment $S^{d,l}$ is included in a given region $\mathcal{R}^{c,k}$. We execute this task in four steps: (1) set the width (thickness) $W_{\mathcal{R}}$ of all regions $\mathcal{R}^{c,k}$ to an arbitrarily small value; (2) define function $g^{c,k}(x)$ as the linear curve crossing both endpoints $p_s^{c,k}$ and $p_e^{c,k}$ of segment $S^{c,k}$; (3) compute $\forall\, h \in \{s, e\}$ the quantity $d_h^{k,l}$, which is the normal distance from point $p_h^{d,l}$ to line $g^{c,k}(x)$; (4) consider that $p_h^{d,l} \in \mathcal{R}^{c,k}$ only if $d_h^{k,l} < t_{roi}$, where $t_{roi} = \frac{W_{\mathcal{R}}}{2}$ is the parameter controlling the width $W_{\mathcal{R}}$. We eventually express the developed filtering condition equivalent to Eq. (7) in Eq. (8), where quantity $d_h^{k,l}$ is computed using Eqs. (9) and (10).

$$S^{d,l} \in S^e \text{ if } \exists k\text{: } \left( \left( d_s^{k,l} \leq t_{roi} \right) \wedge \left( d_e^{k,l} \leq t_{roi} \right) \right) = 1 \quad (8)$$

$$d_h^{k,l} = \frac{\left| g^{c,k}\left(x_h^{d,l}\right) - y_h^{d,l} \right|}{\sqrt{1 + \left(\alpha^{c,k}\right)^2}} \quad \forall h \in \{s, e\} = \frac{\left| \alpha^{c,k} \times x_h^{d,l} + \beta^{c,k} - y_h^{d,l} \right|}{\sqrt{1 + \left(\alpha^{c,k}\right)^2}} \quad (9)$$

$$\beta^{c,k} = y_s^{c,k} - \alpha^{c,k} \times x_s^{c,k} = y_e^{c,k} - \alpha^{c,k} \times x_e^{c,k} \quad (10)$$

where $\alpha^{c,k}$ and $\beta^{c,k}$ correspond to the slope and intercept of curve $g^{c,k}(x)$, respectively. Note that $\alpha^{c,k}$ is equal to the slope of the $k$-th segment $S^{c,k}$ and can be obtained from vector $\alpha^a$.

*G. Vectorizing the Region of Interest Filter (ROIF)*

We present in this Section the details to vectorize the filtering condition we developed in Eq. (8). We define matrix $\mathcal{D}_h$ to contain $\forall\,(k, l)$ the quantities $d_h^{k,l}$. We illustrate $\mathcal{D}_h$ in Eq. (11) and compute it using Eq. (12).

$$\mathcal{D}_h = \begin{bmatrix} d_h^{1,1} & d_h^{1,2} & \cdots & d_h^{1,N^d} \\ d_h^{2,1} & d_h^{2,2} & \cdots & d_h^{2,N^d} \\ \vdots & \vdots & \ddots & \vdots \\ d_h^{N^c,1} & d_h^{N^c,2} & \cdots & d_h^{N^c,N^d} \end{bmatrix} = (d^{k,l}) \in \mathbb{R}^{N^c \times N^d} \quad (11)$$

$$\mathcal{D}_h = \left| \alpha^c \times (x_h^d)^T + B^c - Y_h^d \right| \oslash \left( \left(1^c + (\alpha^c)^{\odot 2}\right)^{\odot \frac{1}{2}} \times (1^d)^T \right) \quad (12)$$

where $(.)^T$ denotes vector transpose, $B^c \in \mathbb{R}^{N^c \times N^d}$ contains intercept values of all functions of set $g^c = \{g^{c,k}\}_{k=1:N^c}$, $Y_h^d \in \mathbb{R}^{N^c \times N^d}$ contains $y$ Cartesian coordinates of all endpoints in set $p_h^d = \{p_h^{d,l}\}_{l=1:N^d}$, and $1^u = [1, 1, \ldots, 1]^T \in \mathbb{R}^{N^u \times 1}$ is an all-ones vector. We provide in what follows the details for getting $\alpha^c$, $B^c$, and $Y_h^d$. We establish $\alpha^c$ by indexing vector $\alpha^a$ using vector $I_{b,a}$, which produces vector $\alpha^b$. We index the latter using vector $I_{c,b}$, producing thus desired slopes $\alpha^c$. We already established both vectors $I_{b,a}$ and $I_{c,b}$ in Section III.D.2. We show elements of $B^c$ in Eq. (13) and compute them by calculating vector $\beta^c$ according to Eq. (14) and broadcasting the result to $N^d$ columns, as shown in Eq. (15). We show elements of matrix $Y_h^d$ in Eq. (16) and compute them using Eq. (17).

$$B^c = \begin{bmatrix} \beta^{c,1} & \beta^{c,1} & \cdots & \beta^{c,1} \\ \beta^{c,2} & \beta^{c,2} & \cdots & \beta^{c,2} \\ \vdots & \vdots & \ddots & \vdots \\ \beta^{c,N^c} & \beta^{c,N^c} & \cdots & \beta^{c,N^c} \end{bmatrix} \in \mathbb{R}^{N^c \times N^d} \quad (13)$$

$$\beta^c = \left[\beta^{c,1}, \beta^{c,2}, \ldots, \beta^{c,N^c}\right] = y_h^c - (\alpha^c \odot x_h^c) \in \mathbb{R}^{N^c \times 1} \quad (14)$$

$$B^c = \beta^c \times (1^d)^T \quad (15)$$

$$Y_h^d = \begin{bmatrix} y_h^{d,1} & y_h^{d,1} & \cdots & y_h^{d,N^d} \\ y_h^{d,1} & y_h^{d,2} & \cdots & y_h^{d,N^d} \\ \vdots & \vdots & \ddots & \vdots \\ y_h^{d,1} & y_h^{d,2} & \cdots & y_h^{d,N^d} \end{bmatrix} \in \mathbb{R}^{N^c \times N^d} \quad (16)$$

$$Y_h^d = 1^c \times \left(y_h^d\right)^T \in \mathbb{R}^{N^c \times N^d} \quad (17)$$

To exploit the distances in matrix $\mathcal{D}_h$, we define $Q_h$ as in Eq. (18) and compute it using Eq. (19).

$$Q_h = \begin{bmatrix} q_h^{1,1} & q_h^{1,2} & \cdots & q_h^{1,N^d} \\ q_h^{2,1} & q_h^{2,2} & \cdots & q_h^{2,N^d} \\ \vdots & \vdots & \ddots & \vdots \\ q_h^{N^c,1} & q_h^{N^c,2} & \cdots & q_h^{N^c,N^d} \end{bmatrix} = \left(q_h^{k,l}\right) \in \mathbb{R}^{N^c \times N^d} \tag{18}$$

$$Q_h = \mathcal{D}_h \le T_{roi} \in \mathbb{R}^{N^c \times N^d} \quad \forall h \in \{s, e\} \tag{19}$$

where $q_h^{k,l}$ is a Boolean scalar whose truth indicates that endpoint $p_h^{d,l} \in \mathcal{R}^{c,k}$ and $T_{roi} \in \mathbb{R}^{N^c \times N^d}$ is a matrix whose all elements are equal to $t_{roi}$ (Eq. (8)). The comparison performed in Eq. (19) is equivalent to that of Eq. (8). We further process matrix $Q_h$ using Eq. (20).

$$q^d = \breve{\vee}\,(Q_s \,\tilde{\wedge}\, Q_e) \in \mathbb{R}^{1 \times N^d} = \left[q^{d,1}, q^{d,2}, \dots q^{d,N^d}\right] \tag{20}$$

where $\breve{\vee}\,(.)$ operator computes the logical *or* along the vertical axis of matrix[9] $Q_h$, $\tilde{\wedge}$ computes the element-wise logical and, $q^{d,l}$ is a logical Boolean whose truth indicates that segment $S^{d,l}$ is encompassed by at least one region of interest $\mathcal{R}^{c,k}$. We note that the operator used in Eq. (20) reflects the operator in Eqs. (7) and (8), whereas the vertical or operator in Eq. (20) reflects the $\exists$ operator in Eqs. (7) and (8). Thus, Eq. (8), equivalent to Eq. (7), becomes equivalent to Eq. (21). Subsequently, we create vector $I_{e,d}$ to contain indices where elements of vector $q^d$ (Eq. (20)) are true Booleans. Finally, we establish set $S^e$ by indexing $S^d$ using $I_{e,d}$. We compute the final filtered set of segments output by the ROIF (Stage 3 in Fig. 5) using Eq. (22). Fig. 6(b) and (f) show an example of the original set of segments $S^a$ and the corresponding filtered set $S^f$, respectively.

$$S^{d,l} \in S^e \ if \ q^{d,l} = 1 \tag{21}$$

$$S^f = S^c \cup S^e \tag{22}$$

### H. Segments to Edge Pixels (STEP)

To get filtered edge points, we must find the coordinates of all pixels along each segment in set $S^f$. The number of pixels to locate for each segment $S^{f,n}$ is equal to its length $L^{f,n}$. Thus, we represent the coordinates of pixels along segment $S^{f,n}$ as in Eq. (23) and compute them using Eq. (24).

$$\begin{aligned} x^n &= \left[x^{n,1}, x^{n,2}, \dots x^{n,L^{f,n}}\right]^T \in \mathbb{R}^{L^{f,n} \times 1} \\ y^n &= \left[y^{n,1}, y^{n,2}, \dots y^{n,L^{f,n}}\right]^T \in \mathbb{R}^{L^{f,n} \times 1} \end{aligned} \tag{23}$$

$$\begin{aligned} x^n &= \frac{x_e^{f,n} - x_s^{f,n}}{L^{f,n} - 1} \odot z^n + x_s^{f,n} \\ y^n &= \frac{y_e^{f,n} - y_s^{f,n}}{L^{f,n} - 1} \odot z^n + y_s^{f,n} \end{aligned} \tag{24}$$

where $z^n = [0, 1, 2, \dots, L^{f,n} - 1]^T \in \mathbb{R}^{L^{f,n} \times 1}$. Because detected segments do not have the same length, we cannot further vectorize Eq. (24) as we did with previous equations. Therefore, we iterate over all segments of $S^f$. At each iteration, we compute Eq. (24) and append the result as expressed in Eq. (25).

$$\begin{aligned} x^{out} &:= x^{out} \frown x^n; \forall n \in \{1, 2, \dots, N^f - 1, N^f\} \\ y^{out} &:= y^{out} \frown y^n; \forall n \in \{1, 2, \dots, N^f - 1, N^f\} \end{aligned} \tag{25}$$

where $:=$ represents value assignment, $\frown$ denotes vector appending, and $x^{out}$ and $y^{out}$ are vectors that would contain Cartesian coordinates of pixels along all segments in set $S^f$. Both $x^{out}$ and $y^{out}$ are initialized to an empty vector at the first iteration. Although this process is not fully vectorized, the effect on real-time performance is insignificant as the portion of survived segments in $S^f$ is tiny compared to other sets. We will see in Section IV that the execution speed achieved is satisfying.

Next, we use vectors $x^{out}$ and $y^{out}$ to establish an edge map image $E'(x, y)$ (Stage 4 in Fig. 5) corresponding to the down-sampled image $I'_{red}$. Fitting the horizon line on $E'$ works well but establishing horizon parameters corresponding to the original frame size requires scaling up the horizon position $Y'$ corresponding to $E'$: $Y = Y' \times \frac{1}{\kappa}$. This affects the detected line because we amplify the error corresponding to $Y'$ by $\frac{1}{\kappa}$. Scaling up parameters of the horizon is necessary not only for comparison with Ground Truth (GT) position $Y^{GT}$ but also for subsequent applications. For instance, the computation of transformation matrices involved in stabilizing original video frames is directly related to horizon parameters $\{Y, \phi\}$ [9, 20, 21]. Therefore, we infer the horizon line by relocating edge points of $E'$ on an edge map $E$ (Stage 4 in Fig. 5) with the original frame size. Thus, we significantly mitigate the amplified error and leverage image downsizing. We compute $E$ by upsampling $E'$ to the original frame size using a bilinear interpolation method, which outputs grayscale image $E''$ with thicker edges. We shrink the thickness of the latter to one pixel using the non-maximum suppression method, followed by a single-thresholding of $E''$ with $E_{th} = 254$. This outputs image $E$, which contains relocated edge points. Fig. 8 shows the edge map $E$ corresponding to challenging conditions, such as weak horizon edges, coastal boundaries, and highly textured regions induced by sea waves, glints, and clouds.

[9] In other words, $\breve{\vee}$ performs the logical *or* operation on elements of each column

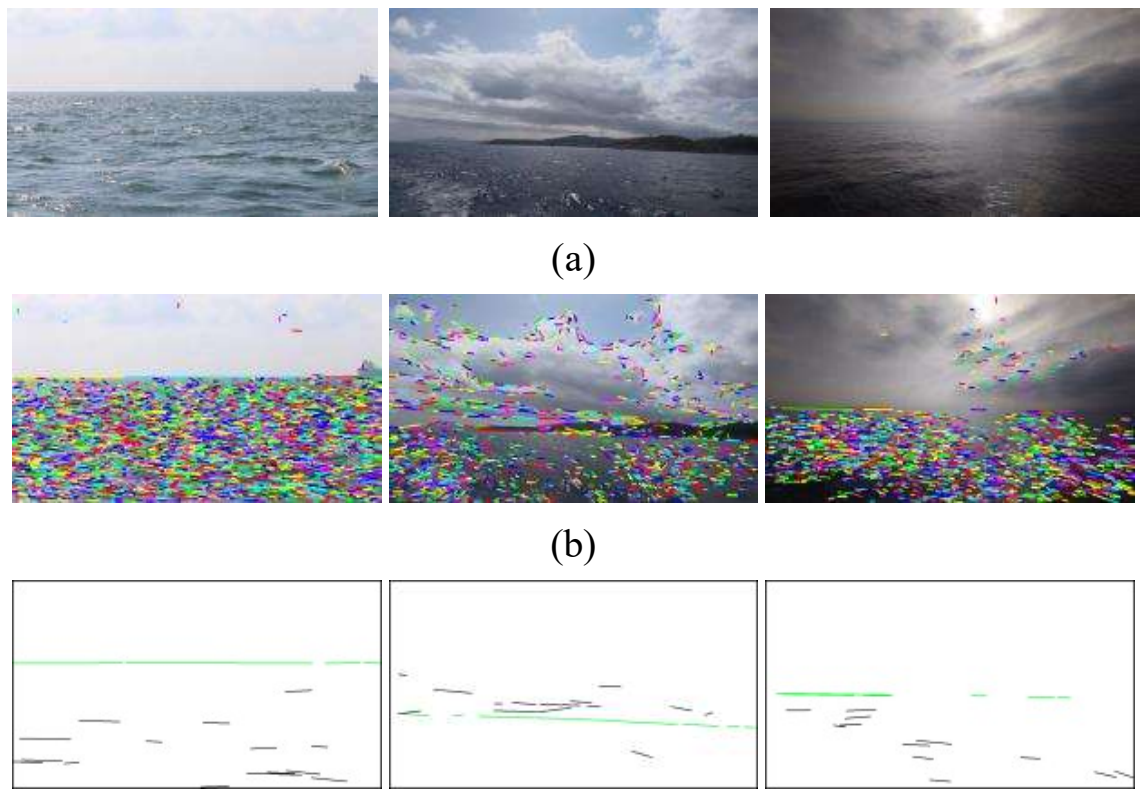

Fig. 8. Illustration of the output edge map $E$. (a) original images, (b) all detected line segments, (c) the output edge map $E$. For visualization convenience, the binary edges are dilated and inverted, and the horizon edges are colored in green.

The large number of segments $S^a$ in Fig. 8(b) indicates the high amount of noise. Fig. 8(c) demonstrates that the high image clutter is significantly suppressed while favoring horizon edges to form the most prominent line. Edges of the horizon line corresponding to the right image of Fig. 8(a) are dramatically affected by the smooth color transition between the sea and sky. However, the corresponding edge map in Fig. 8(c) demonstrates that the filtering stages kept enough horizon edges for accurate detection. This will be further demonstrated qualitatively and quantitatively in Section IV.

### *I. Line Inference*

We infer the horizon line in three major steps (Stage 5 in Fig. 5). The Outlier Handler Module (OHM) verifies that the global maximum of the accumulated Hough space, denoted as $\{Y^{\mathcal{H}}, \phi^{\mathcal{H}}\}$, is not an outlier. We define outlier lines as false-positive horizons persisting over a very brief period. This issue occurs on other algorithms we tested in this paper as well. In our case, outlier lines appear primarily due to noisy segments of the sea region that occasionally line up to survive the ROIF. The OHM considers $\{Y^{\mathcal{H}}, \phi^{\mathcal{H}}\}$ as an outlier line if the condition in Eq. (26) is satisfied.

$$(|Y^{\mathcal{H}} - Y^{prv}| > \Delta Y^{th}) \vee (|\phi^{\mathcal{H}} - \phi^{prv}| > \Delta\phi^{th}) = 1 \quad (26)$$

where $\vee$ is the logical *or*, $\{Y^{prv}, \phi^{prv}\}$ are parameters of the most recent horizon line, and $\{\Delta Y^{th}, \Delta \phi^{th}\}$ are scalar thresholds. If $\{Y^{\mathcal{H}}, \phi^{\mathcal{H}}\}$ is not an outlier, we consider it a coarse estimation of the horizon line: $\{Y^{\mathcal{H}}, \phi^{\mathcal{H}}\} = \{Y^{crs}, \phi^{crs}\}$. Otherwise, if $\{Y^{\mathcal{H}}, \phi^{\mathcal{H}}\}$ is an outlier, the OHM will consider that the coarse horizon line $\{Y^{crs}, \phi^{crs}\}$ is one of the longest M Hough lines satisfying Eq. (26). If multiple lines satisfy such a condition, we select the line corresponding to the minimum value $|Y^{\mathcal{H}} - Y^{prv}|$. To refine the coarse line $\{Y^{crs}, \phi^{crs}\}$, we follow Jeong *et al.* [39] by applying the least-squares fitting algorithm on *inlier edge points*, i.e., edge points that voted on $\{Y^{crs}, \phi^{crs}\}$.

Ettinger *et al.* [43] pointed out that using previous detections of the horizon line to infer the horizon may get the algorithm into a failure state, concretely. If the algorithm detects a faulty line, it will incorrectly update the parameters of the most recent horizon line $\{Y^{prv}, \phi^{prv}\}$. Therefore, the outlier condition in Eq. (26) becomes useless if the difference between that faulty line and the true horizon exceeds a certain threshold. When this issue occurs, the faulty horizon is unlikely to be persistently detected on subsequent frames, either due to the changing sea noise or the linear property of the horizon that would compete with that faulty line. Thus, the outlier condition triggers on a significant number of consecutive frames. The OHM counts this number, denoted as $N_{outs}$, and compares it to a threshold $N_{outs}^{th}$. If $N_{outs} > N_{outs}^{th}$, the OHM considers that the algorithm is in a failure state. To get the algorithm out of this state, the OHM avoids finding a substitute line and directly refines the longest Hough line $\{Y^{\mathcal{H}}, \phi^{\mathcal{H}}\}$ using the least-squares fitting. This process quickly converges to correctly updating the parameters $\{Y^{prv}, \phi^{prv}\}$. The reader may question the usefulness of the OHM as it seems to create the same problem that it solves. From one hand, the OHM is supposed to correct outlier horizons, which are fundamentally incorrect detections. On the one hand, the OHM may get the algorithm into a failure state that may lead to incorrect detections again. The OHM is effective because the incorrect detections caused by failure states are negligible compared to the incorrect detections corresponding to outlier lines. We quantitatively justify the effectiveness of the OHM in Section IV.F by ablating it from the overall algorithm.

## IV. Results and Discussion

### *A. Hardware and Software*

We compare our algorithm with an extensive set of eight state-of-the-art algorithms: [14–17, 29, 30, 37, 39]. The original authors of [16, 37] provided their MATLAB implementation. We implemented all the remaining algorithms [14, 15, 17, 29, 30, 39], including ours, on Python 3.8, utilizing the packages Numpy, OpenCV, and TensorFlow [53–55]. Since the eager execution of TensorFlow is slow and inefficient, we built the CNN model of [29] as a graph instance, which allows faster and optimal computations using the graph execution mode. We ran each compared algorithm, one at a time, on a computer with 16GB of RAM and one Intel® Core™ i5-6300U CPU @ 2.40GHz, ensuring a fair comparison by halting all tasks that may use the computer's CPU during each algorithm's execution.

We describe in what follows important considerations regarding the algorithms we implemented. Since Dong *et al.* [14] operates on infrared images and the authors do not specify the number of iterations used in their RANSAC line fitting, we used a grayscale input image computed by averaging the RGB channels and ran the RANSAC algorithm on all possible pairs of segment endpoints. We motivate the last modification by the limited number of possible pairs, which results from the small number of survived segments and the constraint of the pairs to belong to the same line segment. In the same context of fitting the horizon line, [17] use Dynamic Programming (DP) to infer the non-linear horizon of wild mountainous scenes. Since the horizon in our case is a straight line, we substituted the DP inference using the most accurate horizon line fitting [29, 39]. While the algorithm proposed in [17] was not originally intended for maritime images, we included it in our experimental comparison due to its reported capability to extract weak horizon edges from scenes captured in various wild environments.

### B. Parameters of Our Algorithm

For generalization purposes, we did not perform any quantitative optimization study to fix the parameters of our algorithm. Instead, we selected their values based on our knowledge of their effect and our visual analysis of intermediate results (i.e., the output of different stages and modules) on a representative set of 85 images. All the quantitative and qualitative results reported in this paper use the exact values in Table III.

TABLE III. VALUES OF OUR METHOD'S PARAMETERS

| Stages | Parameter values |
|---|---|
| Stage 1 | Down-sampling factor: $\kappa = 0.6$; LSD parameters: Sobel kernel = 3 × 3, Canny's upper and lower thresholds: 25 and 45. |
| Stage 2 | $\alpha_{th} = 0.57$, $N^c = 15$; $N^d = 200$ |
| Stage 3 | $t_{roi} = 2$ pixels |
| Stage 4 | No parameters |
| Stage 5 | Hough space $H(\rho, \theta)$: $\Delta\rho = 2$ pixels, $\Delta\theta = 1°$; OHM: $\Delta Y^{th} = 2.6\,\% \times$ Image Height $= 50$ pixels, $\Delta\phi^{th} = 2°$, $M = 2$, $N_{outs}^{th} = 5$. |

### C. Comparison Metrics

Following the sea horizon literature [6, 12, 16, 29, 39], we parameterize the detected horizon line using the position $Y$ and tilt $\phi$ we previously exposed in Fig. 2. We quantify the detection error by computing the positional error and angular error, as in Eqs. (27) and (28):

$$Y^{\epsilon} = |Y - Y^{GT}| \quad (27)$$

$$\phi^{\epsilon} = |\phi - \phi^{GT}| \quad (28)$$

where the pairs $\{Y^{GT}, \phi^{GT}\}$ and $\{Y, \phi\}$ correspond to parameters of the GT (Ground Truth) and the detected horizon, respectively. Additionally, we introduce a new *composite* metric $\widetilde{C^{\epsilon}}$, which we justify in Section IV.D, for measuring the performance of horizon detection algorithms, merging both $Y^{\epsilon}$ and $\phi^{\epsilon}$ into one value, as shown in Eq. (29).

$$\widetilde{C^{\epsilon}} = 100 \times \sqrt{W_Y \times \widetilde{Y^{\epsilon}} + W_{\phi} \times \widetilde{\phi^{\epsilon}}} \quad (29)$$

where $\widetilde{Y^{\epsilon}} \in [0,1]$ and $\widetilde{\phi^{\epsilon}} \in [0,1]$ correspond to normalization of errors $Y^{\epsilon}$ and $\phi^{\epsilon}$, respectively. $W_Y$ and $W_{\phi}$ represent weighting values summing up to 1, ensuring that the quantity $\widetilde{C^{\epsilon}}$ is in the range [0, 100%] (worst at 100%, best at 0%). We assume that positional and angular errors are equally important and set $W_Y = W_{\phi} = 0.5$. Eventually, we use these three errors (i.e., $Y^{\epsilon}$, $\phi^{\epsilon}$, and $\widetilde{C^{\epsilon}}$) computed for each dataset to extract the following six statistical metrics: $\mu$, $\sigma$, Q25, Q50, Q75, and Q95, where $\mu$ and $\sigma$ correspond to the mean and standard deviation, respectively, and QP is the P-th percentile. To quantify the real-time performance, we measure the mean processing time of all algorithms on each dataset. We also report the visual results of each algorithm on six representative maritime images.

### D. Justification of the New Composite Metric

Researchers have been using the positional error $Y^{\epsilon}$ and angular error $\phi^{\epsilon}$ as comparison metrics due to their interpretability. Sometimes, one of the errors (e.g., the angular error $\phi^{\epsilon}$ ) may be very low. However, its corresponding positional error may be high, as seen in Fig. 9, where the strong faulty line is in parallel with the true horizon line. Therefore, to enhance the experimental comparison, we introduce a third metric that we derived from the semantic line detection literature [56]. In this context, Zhao *et al.* [56] explained the drawbacks of various metrics and concluded by introducing a new metric that merges both angular and positional error, as shown in Eq. (30).

$$S = (S_{\theta} \times S_d)^2 \quad (30)$$

where $S_{\theta}$ and $S_d$ measure the normalized angular error and positional error. Thus, it is tempting to derive a similar metric from Eq. (30), as in Eq. (31).

$$S' = \left(\widetilde{Y^{\epsilon}} \times \widetilde{\phi^{\epsilon}}\right)^2 \quad (31)$$

where the notation $\widetilde{(.)}$ normalizes the value into the range [0, 1].

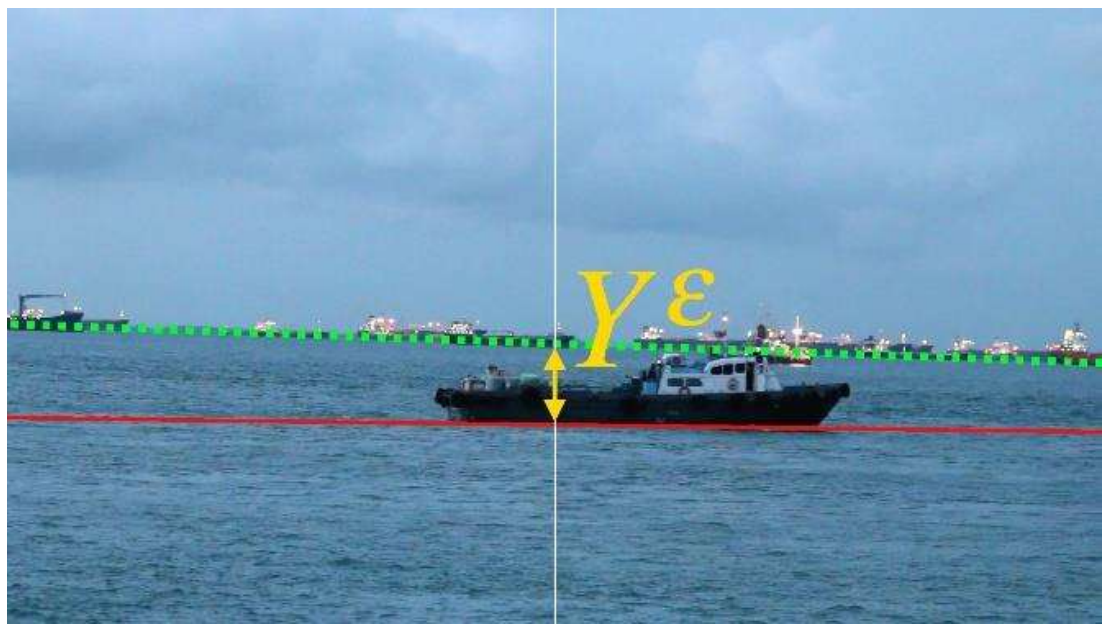

Fig. 9. An incorrect detection (in red) by a state-of-the-art algorithm caused by the strong faulty line induced by the ship edges. This line being almost in parallel with the true horizon (in dotted green) significantly reduces the angular error $\phi^{\epsilon}$, but the corresponding positional error $Y^{\epsilon}$ is high.

After applying the metric in Eq. (31) on state-of-the-art algorithms, we observed a significant bias of the statistical errors towards zero, falsely indicating a good performance even when a given algorithm performs poorly. This bias is due to the multiplication operation in Eq. (31), which reduces high errors if one of the multiplied terms is low[10]. Therefore, we devised the composite error $\widetilde{C^{\epsilon}}$ in Eq. (29), which avoids the bias caused by multiplying errors. This is achieved by accumulating the errors through a weighted sum. Additionally, square rooting makes the metric much more sensitive and discriminative as the composite error approaches zero.

### *E. State-of-the-Art Comparison: Result and Discussion*

Tables IV to XII quantitatively compare the statistical errors of all nine algorithms on the SMD-onboard (Tables IV, V, and VI), SMD-onshore (Tables VII, VIII, and IX), and the TMD (Tables X, XI, and XII). Fig. 10 quantitatively compares all algorithms on six challenging images. The statistical errors obtained for the SMD-Onboard and SMD-Onshore show that our algorithm and Jeong *et al.*'s method [39] perform significantly better. On the SMD-Onboard dataset (Tables IV, V, and VI), our method performed the best in 12 statistical metrics (out of 18), while Jeong *et al.*'s method [39] scored the best on the remaining six metrics. We expected this particular result for [39] because its multi-scale median filter is highly effective against various sea clutter when the horizon has a prominent length and edges, which is the case in most SMD-onboard videos.

Compared to the SMD-Onboard dataset, the statistical results we obtained for the SMD-Onshore (Tables VII, VIII, and IX) show a performance decline for [15, 16, 29, 39] and our method. Our decline is likely because the horizon in SMD-onshore videos is relatively shorter due to ship occlusion and mixed with additional clutter, such as linear wakes and ships' contours. This interpretation may apply, at least to some extent, to the other methods. Nevertheless, these results in Tables VII, VIII, and IX, show that our method performs remarkably better in 17 (out of 18) statistical metrics while scoring a very competitive value on the remaining metric σ in Table VII. Tables X, XI, and XII, show that our algorithm achieves the lowest errors on all the 18 statistical metrics when tested on the TMD, and the error gap between our method and all eight state-of-the-art algorithms is remarkably high. As the TMD introduces multiple new conditions, including image scenes with low horizon contrast, the last result is a strong indicator of the significant robustness of our method against weak and degraded horizon edges. While Dong *et al.* [14] report high robustness against infrared weak horizon edges, Tables X, XI, and XII, and Fig. 10(d) provide quantitative and qualitative evidence for the unsuitability of their infrared filter for RGB images. The quantitative errors corresponding to SMD videos (Tables IV, V, VI, VII, VIII, and IX) provide further evidence of the significance of customizing the algorithm according to the nature of the scene and image. Specifically, the top two performing algorithms for a particular dataset, excluding our own, have been specifically designed for maritime visible-range images [15, 29, 37, 39]. In contrast, the least performing methods have been originally designed for different types of images (e.g., infrared images, wild images) [14, 17] or use general image processing techniques [17, 30].

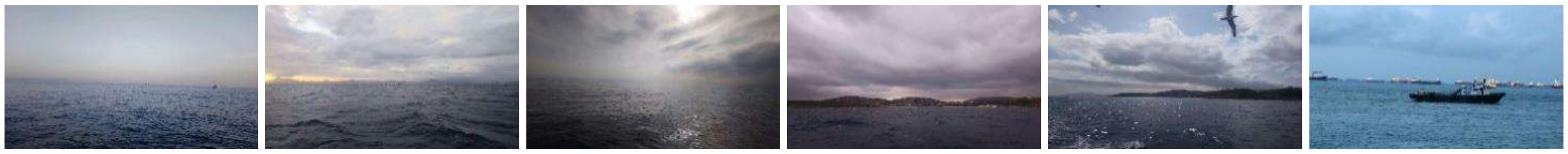

(a) Original images

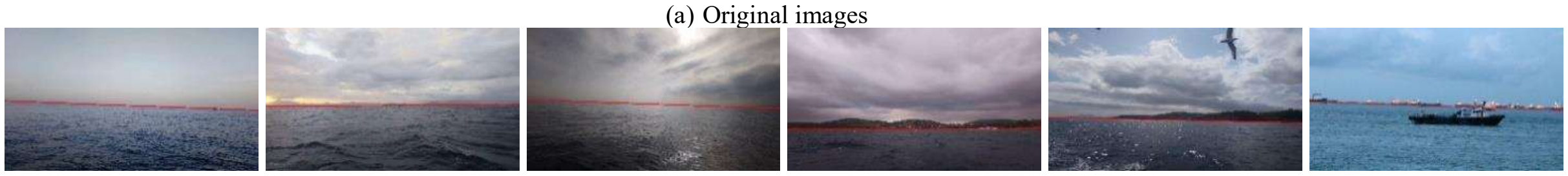

(b) Results on samples of Fig. 10(a) (Ours)

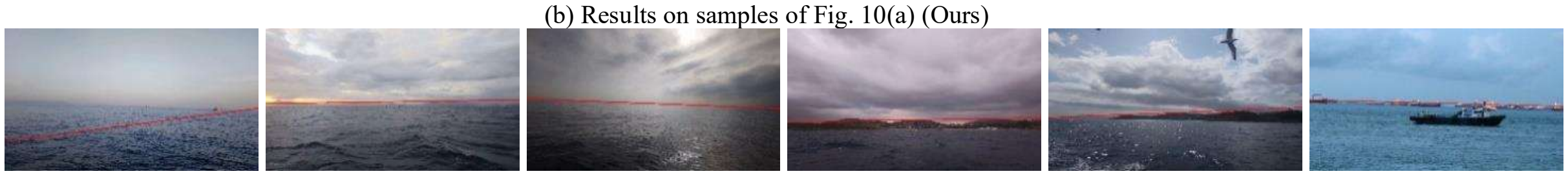

(c) Results on samples of Fig. 10(a) (Li *et al.* [15])

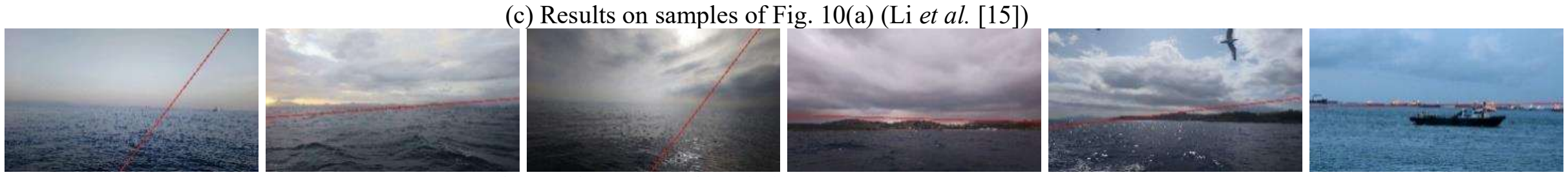

(d) Results on samples of Fig. 10(a) (Dong *et al.* [14])

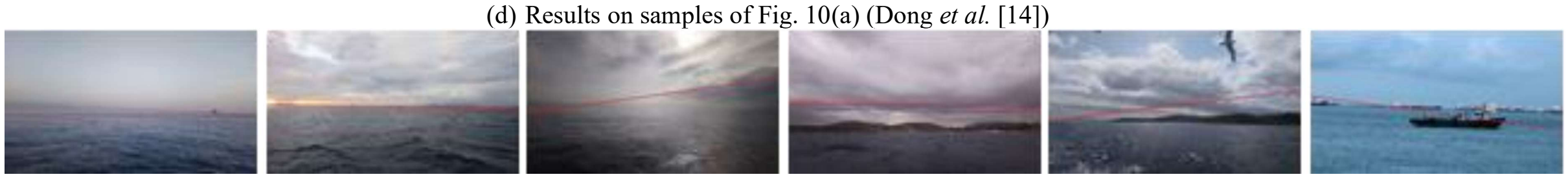

(e) Results on samples of Fig. 10(a) (Liang and Liang [16])

[10] We emphasize that our work does not suggest a bias in the experiments of Zhao *et al.* [56] since their method for computing positional and angular errors differs from ours.

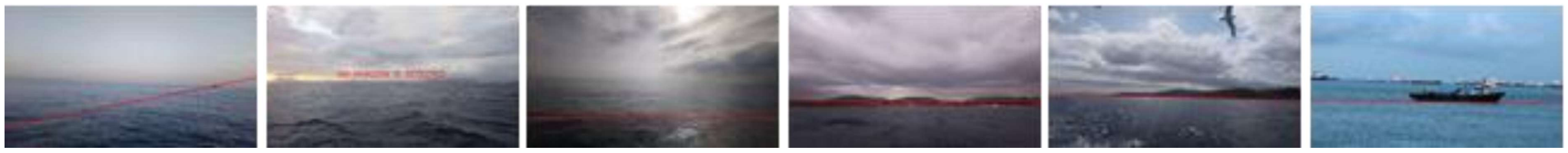

(f) Results on samples of Fig. 10(a) (Jeong *et al.* [39])

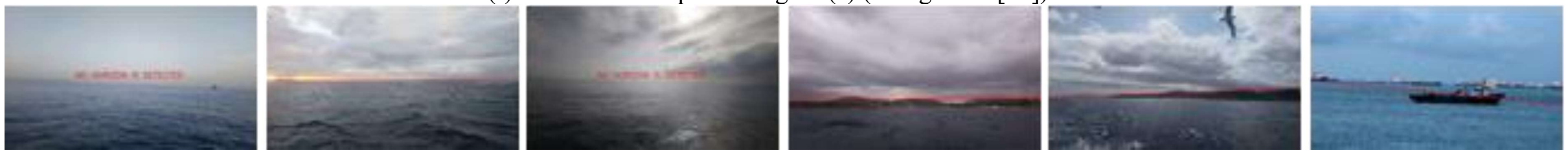

(g) Results on samples of Fig. 10(a) (Jeong *et al.* [29])

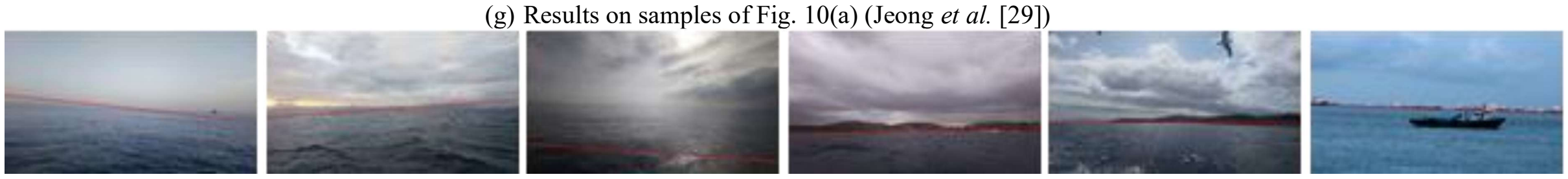

(h) Results on samples of Fig. 10(a) (Prasad *et al.* [37])

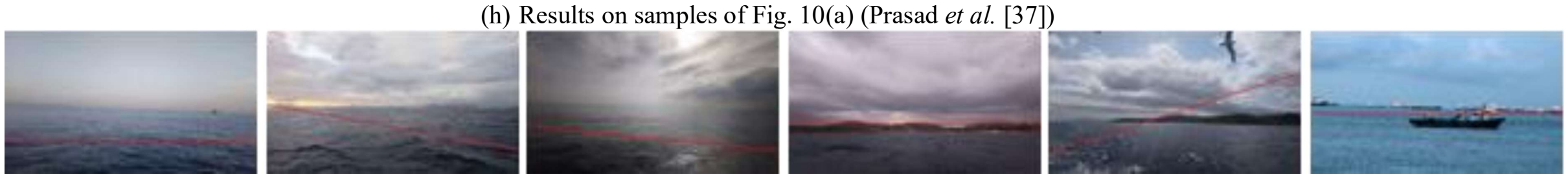

(i) Results on samples of Fig. 10(a) (Gershikov *et al.* [30])

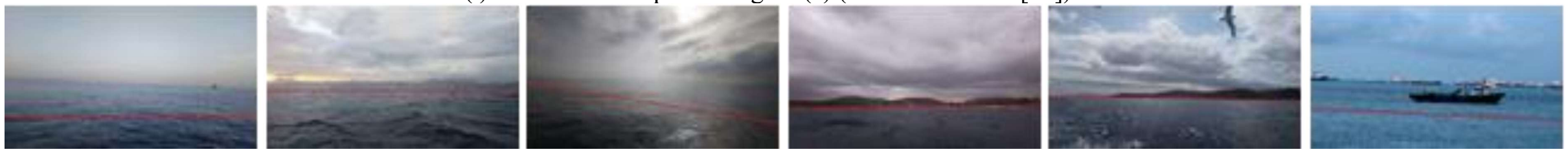

(j) Results on samples of Fig. 10(a) (Lie *et al.* [17])

Fig. 10. Qualitative comparison on six image samples.

In Fig. 10, Figs. 10(a) to (j) visually compare the algorithms' output on six image samples. By comparing the first three of images (starting from the left image) in Fig. 10(b) with state-of-the-art, we can observe that our algorithm is the only one that reliably and accurately works under very weak horizon edges. Moreover, the last three images of Fig. 10(b) shows accurate results in other challenging conditions, including strong and connected coastal boundaries, sun glints, textured clouds, and ships' contours. Although the horizon in coastal images may be a prominent strong line (e.g., the 5th image in Fig. 10(a)), several state-of-the-art algorithms failed, which is in some cases due to the strong sky-coast edges. We observed that such edges do not affect our algorithm due to the instability of their gradient angles, which constrains the length of corresponding line segments and increases their suppression rate at Stage 2 (Fig. 5).

The fastest algorithm is [30], while the slowest is [37]. From the perspective of computational speed, our algorithm is not the fastest as it takes, on average, 1.78 times longer than the fastest algorithm. However, on the onboard and onshore datasets, we maintain a very competitive speed with the next fastest four methods [14, 16, 17, 39]. Additionally, our algorithm is approximately four times faster than the fastest algorithm among [15, 29, 37]. On the TMD, our algorithm is the fastest after [30]. This difference in speed compared to the other two datasets is due to the lower contrast of images, which means the LSD grows fewer line segments, hence less processing time. It is remarkable that [14], which also grows and filters line segments, has a more stable execution time than our method. This stability is likely because [14] uses an adaptive threshold that stabilizes the number of grown segments while we use fixed low thresholds on all images to ensure the extraction of weak horizon edges. The computational speed is an essential factor, but the real-time performance also involves error performance as well. While our method is 1.78 slower than the fastest method [30], the latter has for instance, a mean positional error $Y^{\epsilon}$ that's at least 21 times higher than ours. While we maintain a highly competitive speed against the most accurate methods, our error performance surpasses both the fastest [14, 16, 17, 30, 39] and the slowest algorithms [15, 29, 37] with a significant margin.

TABLE IV. Positional Error Statistics (Positional Error $Y^{\epsilon}$ in Pixels) for the SMD-Onboard

| Algorithm | $\mu$ | $\sigma$ | Q25 | Q50 | Q95 | Q95 |
|---|---|---|---|---|---|---|
| Ours | **1.95** | **1.59** | 0.84 | 1.71 | 2.71 | **4.57** |
| Li [15] | 34.64 | 73.96 | 14.35 | 22.72 | 30.84 | 71.49 |
| Dong [14] | 116.95 | 689.88 | 3.78 | 17.87 | 5.34 | 292.55 |
| Liang [16] | 17.77 | 46.93 | 0.90 | 2.13 | 7.65 | 93.43 |
| Jeong [39] | 3.84 | 33.09 | **0.68** | **1.43** | **2.44** | 4.86 |
| Jeong [29] | 13.71 | 214.61 | 0.78 | 1.63 | 2.89 | 10.13 |
| Prasad [37] | 210.50 | 212.81 | 2.51 | 198.68 | 386.00 | 564.98 |
| Gershikov [30] | 301.44 | 145.02 | 186.67 | 334.50 | 410.28 | 510.67 |
| Lie [17] | 210.67 | 129.62 | 115.89 | 214.81 | 297.31 | 429.40 |

TABLE V. Angular Error Statistics (Angular Error $\Phi^{\epsilon}$ in Degrees) for the SMD-Onboard

| Algorithm | $\mu$ | $\sigma$ | Q25 | Q50 | Q95 | Q95 |
|---|---|---|---|---|---|---|
| Ours | **0.16** | **0.13** | **0.060** | **0.13** | **0.23** | **0.41** |
| Li [15] | 0.85 | 1.46 | 0.37 | 0.64 | 1.0 | 1.70 |
| Dong [14] | 15.78 | 21.93 | 0.34 | 5.43 | 21.43 | 65.29 |
| Liang [16] | 4.71 | 3.38 | 2.53 | 4.27 | 6.28 | 10.09 |
| Jeong [39] | 0.26 | 1.61 | 0.065 | 0.144 | 0.25 | 0.50 |
| Jeong [29] | 0.58 | 4.14 | 0.066 | 0.147 | 0.26 | 0.66 |
| Prasad [37] | 2.65 | 2.40 | 0.84 | 2.00 | 3.74 | 7.50 |
| Gershikov [30] | 6.55 | 6.85 | 1.60 | 3.84 | 9.17 | 22.10 |
| Lie [17] | 2.26 | 1.95 | 0.71 | 1.94 | 3.26 | 6.00 |

TABLE VI. COMPOSITE ERROR STATISTICS (NORMALIZED COMPOSITE ERROR $C^c$ IN %) FOR THE SMD-ONBOARD

| Algorithm | *μ* | *σ* | Q25 | Q50 | Q95 | Q95 |
|---|---|---|---|---|---|---|
| Ours | **3.51** | **1.11** | 2.75 | 3.48 | 4.21 | **5.36** |
| Li [15] | 11.83 | 6.64 | 8.9 | 11.1 | 12.99 | 19.63 |
| Dong [14] | 21.62 | 22.64 | 5.48 | 16.56 | 29.32 | 54.27 |
| Liang [16] | 13.2 | 6.25 | 9.33 | 12.19 | 14.98 | 23.76 |
| Jeong [39] | 3.75 | 3.31 | **2.62** | **3.35** | **4.14** | 5.66 |
| Jeong [29] | 4.52 | 7.69 | 2.72 | 3.48 | 4.42 | 10.68 |
| Prasad [37] | 26.85 | 18.09 | 8.12 | 31.32 | 43.21 | 52.3 |
| Gershikov [30] | 38.47 | 9.88 | 33.18 | 41.0 | 45.04 | 51.39 |
| Lie [17] | 29.86 | 12.12 | 24.28 | 32.75 | 38.21 | 45.45 |

TABLE VII. POSITIONAL ERROR STATISTICS (POSITIONAL ERROR $Y^c$ IN PIXELS) FOR THE SMD-ONSHORE

| Algorithm | *μ* | *σ* | Q25 | Q50 | Q95 | Q95 |
|---|---|---|---|---|---|---|
| Ours | **5.66** | **11.20** | **1.20** | **2.56** | **5.15** | **22.62** |
| Li [15] | 50.49 | 97.13 | 7.24 | 17.73 | 48.14 | 246.26 |
| Dong [14] | 160.23 | 791.70 | 17.80 | 40.48 | 99.44 | 481.16 |
| Liang [16] | 31.50 | 49.19 | 2.50 | 17.65 | 33.13 | 151.97 |
| Jeong [39] | 7.86 | 12.56 | 1.68 | 3.61 | 8.60 | 29.62 |
| Jeong [29] | 18.20 | 33.12 | 2.52 | 9.03 | 19.96 | 67.11 |
| Prasad [37] | 12.83 | 37.99 | 1.93 | 3.60 | 7.49 | 43.18 |
| Gershikov [30] | 119.03 | 93.29 | 30.62 | 99.83 | 185.74 | 284.41 |
| Lie [17] | 199.23 | 147.25 | 101.05 | 186.07 | 275.21 | 484.29 |

TABLE VIII. ANGULAR ERROR STATISTICS (ANGULAR ERROR $\Phi^c$ IN DEGREES) FOR THE SMD-ONSHORE

| Algorithm | *μ* | *σ* | Q25 | Q50 | Q95 | Q95 |
|---|---|---|---|---|---|---|
| Ours | **0.22** | 0.25 | **0.05** | **0.14** | **0.33** | **0.66** |
| Li [15] | 1.07 | 2.23 | 0.17 | 0.39 | 1.13 | 4.26 |
| Dong [14] | 27.49 | 25.19 | 5.70 | 20.29 | 52.34 | 78.27 |
| Liang [16] | 3.40 | 4.13 | 0.48 | 1.72 | 4.85 | 12.81 |
| Jeong [39] | 0.28 | **0.22** | 0.12 | 0.24 | 0.37 | 0.72 |
| Jeong [29] | 1.16 | 2.07 | 0.12 | 0.44 | 1.62 | 3.814 |
| Prasad [37] | 2.16 | 2.29 | 0.30 | 1.86 | 3.75 | 6.33 |
| Gershikov [30] | 5.09 | 5.37 | 1.19 | 2.89 | 7.20 | 17.20 |
| Lie [17] | 1.30 | 1.86 | 0.16 | 0.41 | 1.76 | 5.21 |

TABLE IX. COMPOSITE ERROR STATISTICS (NORMALIZED COMPOSITE ERROR $C^c$ IN %) FOR THE SMD-ONSHORE

| Algorithm | *μ* | *σ* | Q25 | Q50 | Q95 | Q95 |
|---|---|---|---|---|---|---|
| Ours | **4.91** | **2.9** | **3.01** | **4.05** | **5.7** | **10.64** |
| Li [15] | 13.06 | 9.65 | 6.67 | 10.36 | 16.41 | 35.81 |
| Dong [14] | 31.65 | 22.44 | 17.2 | 26.7 | 41.72 | 65.03 |
| Liang [16] | 13.04 | 8.4 | 5.28 | 13.66 | 16.68 | 31.48 |
| Jeong [39] | 5.8 | 3.25 | 3.61 | 4.69 | 7.32 | 12.1 |
| Jeong [29] | 8.99 | 6.0 | 4.22 | 8.36 | 11.96 | 19.72 |
| Prasad [37] | 9.34 | 5.67 | 5.41 | 8.42 | 11.59 | 18.37 |
| Gershikov [30] | 24.25 | 10.22 | 14.3 | 25.29 | 32.58 | 38.9 |
| Lie [17] | 28.25 | 12.67 | 21.91 | 29.99 | 36.41 | 47.91 |

TABLE X. POSITIONAL ERROR STATISTICS (POSITIONAL ERROR $Y^c$ IN PIXELS) FOR THE TMD

| Algorithm | *μ* | *σ* | Q25 | Q50 | Q95 | Q95 |
|---|---|---|---|---|---|---|
| Ours | **4.35** | **3.70** | **2.16** | **3.72** | **5.70** | **9.21** |
| Li [15] | 50.32 | 81.28 | 8.83 | 19.21 | 52.85 | 245.61 |
| Dong [14] | 195.01 | 946.18 | 25.23 | 71.40 | 170.32 | 418.30 |
| Liang [16] | 123.57 | 189.78 | 10.31 | 55.66 | 143.41 | 749.97 |
| Jeong [39] | 150.09 | 3785.0 | 2.67 | 5.47 | 28.03 | 225.98 |
| Jeong [29] | 71.83 | 136.67 | 3.20 | 30.14 | 67.32 | 400.95 |
| Prasad [37] | 99.62 | 143.73 | 3.54 | 8.00 | 197.49 | 396.20 |
| Gershikov [30] | 148.15 | 100.10 | 63.37 | 143.38 | 205.94 | 331.05 |
| Lie [17] | 118.43 | 108.17 | 16.46 | 95.17 | 197.25 | 313.10 |

TABLE XI. ANGULAR ERROR STATISTICS (ANGULAR ERROR $\Phi^c$ IN DEGREES) FOR THE TMD

| Algorithm | *μ* | *σ* | Q25 | Q50 | Q95 | Q95 |
|---|---|---|---|---|---|---|
| Ours | **0.23** | **0.26** | **0.07** | **0.15** | **0.32** | **0.63** |
| Li [15] | 1.63 | 2.66 | 0.37 | 0.80 | 1.47 | 7.20 |
| Dong [14] | 18.07 | 21.69 | 3.06 | 9.09 | 22.23 | 67.27 |
| Liang [16] | 4.94 | 4.87 | 1.71 | 3.74 | 6.66 | 13.26 |
| Jeong [39] | 1.81 | 7.42 | 0.12 | 0.33 | 0.66 | 6.90 |
| Jeong [29] | 3.62 | 8.46 | 0.11 | 0.40 | 3.44 | 18.70 |
| Prasad [37] | 4.89 | 4.37 | 1.72 | 3.78 | 7.01 | 13.19 |
| Gershikov [30] | 7.14 | 5.90 | 2.37 | 5.78 | 10.26 | 19.61 |
| Lie [17] | 2.15 | 2.53 | 0.36 | 1.23 | 3.00 | 7.68 |

TABLE XII. COMPOSITE ERROR STATISTICS (NORMALIZED COMPOSITE ERROR $C^c$ IN %) FOR THE TMD

| Algorithm | *μ* | *σ* | Q25 | Q50 | Q95 | Q95 |
|---|---|---|---|---|---|---|
| Ours | **4.84** | **1.81** | **3.6** | **4.65** | **5.84** | **7.64** |
| Li [15] | 13.77 | 9.42 | 7.56 | 10.61 | 17.19 | 36.25 |
| Dong [14] | 30.16 | 22.25 | 18.12 | 26.44 | 38.38 | 59.37 |
| Liang [16] | 22.82 | 13.73 | 12.79 | 19.56 | 29.54 | 59.26 |
| Jeong [39] | 10.9 | 25.03 | 4.05 | 6.01 | 12.35 | 34.69 |
| Jeong [29] | 15.53 | 13.6 | 4.41 | 12.75 | 19.31 | 44.21 |
| Prasad [37] | 20.47 | 13.36 | 10.21 | 15.02 | 32.64 | 44.72 |
| Gershikov [30] | 28.52 | 8.42 | 22.65 | 28.95 | 33.31 | 44.04 |
| Lie [17] | 21.32 | 12.4 | 9.93 | 22.25 | 32.1 | 39.49 |

Table XIII shows the execution time of all methods on each dataset.

TABLE XIII. MEAN COMPUTATIONAL TIME PER FRAME ON THE SMD (ONBOARD AND ONSHORE) AND TMD (IN MILLISECONDS)

| Algorithm | SMD-Onboard | SMD-Onshore | TMD |
|---|---|---|---|
| Ours | 107.6 | 101.4 | 70.3 |
| Li [15] | 486.5 | 426.0 | 418.8 |
| Dong [14] | 87.6 | 87.5 | 88.5 |
| Liang [16] | 73.7 | 79.4 | 74.6 |
| Jeong [39] | 90.6 | 95.7 | 85.4 |
| Jeong [29] | 1000.2 | 685.5 | 748.8 |
| Prasad [37] | 1979.4 | 1946.1 | 1932.6 |
| Gershikov [30] | 50.6 | 53.0 | 52.7 |
| Lie [17] | 141.8 | 101.4 | 111.5 |

Table XIV indicates that the proposed filter, i.e., stages 2, 3, and 4, runs at a blazing speed as it does not exceed 1.8% of the overall computations. Notably, the computational time of the used LSD [51] is at least 62%; the literature on LSD includes better algorithms. For instance, the LSD proposed by Akinlar and Topal [52] is 11 times faster than the LSD [51] we used in this paper. Thus, we can significantly boost the computational speed of our method while enhancing its accuracy just by using an alternative LSD in stage 1 (Fig. 5).

TABLE XIV. THE COMPUTATIONAL TIME RATIO OF COMPONENTS OF OUR ALGORITHM FROM THE MEAN TIME SHOWN IN THE FIRST ROW OF TABLE XIII

| Stage/Component | SMD-Onboard | SMD-Onshore | TMD |
|---|---|---|---|
| LSD [51] | 75.38% | 72.37% | 62.96% |
| Stage 1 | 80.52% | 78.5% | 71.66% |
| Stage 2, 3, and 4 | 0.8% | 1.52% | 1.71% |
| Stage 5 | 18.68% | 19.94% | 26.48% |

### *F. Ablation Study*

In this section, we ablate the key components of our algorithm. The results presented in Table XV demonstrate that the original algorithm, incorporating both the ROIF and OHM, achieves the highest accuracy with the lowest errors in 14 out of 18 statistical metrics. Removing either the ROIF or OHM component increases the errors, confirming the importance of both components.

TABLE XV. THE ERROR STATISTICS COMPUTED AFTER ABLATING THE ROIF AND OHM COMPONENTS. THE RESULTS ARE OBTAINED ON THE COMBINATION OF THE SMD-ONBOARD, SMD-ONSHORE, AND THE TMD, AMOUNTING TO 26,125 FRAMES

| Ablated | | $\mu$ | $\sigma$ | Q25 | Q50 | Q95 | Q95 |
|---|---|---|---|---|---|---|---|
| Positional error $Y^{\epsilon}$ in pixels | None | 4.96 | 9.36 | **1.30** | **2.69** | **4.98** | 20.28 |
| | ROIF | **4.93** | **7.46** | 1.37 | 2.74 | 5.18 | **19.82** |
| | OHM | 7.51 | 28.12 | 1.31 | 2.73 | 5.12 | 22.15 |
| Angular error $\phi^{\epsilon}$ in degrees | None | **0.22** | **0.23** | **0.054** | **0.14** | **0.31** | **0.64** |
| | ROIF | 0.23 | 0.25 | 0.055 | 0.15 | 0.33 | 0.67 |
| | OHM | 0.23 | 0.30 | 0.055 | 0.15 | 0.32 | 0.66 |
| Normalized composite error $C^{\epsilon}$ in % | None | **4.74** | 2.58 | **3.09** | **4.11** | **5.54** | **10.01** |
| | ROIF | 4.80 | **2.52** | 3.13 | 4.12 | 5.68 | 10.02 |
| | OHM | 5.07 | 3.96 | 3.10 | 4.13 | 5.61 | 10.60 |

### *G. Limitations*

We illustrate the limitation of the proposed algorithm in Fig. 11, which occurs due to occasional scenarios depicting faulty lines with stable gradient orientations. The limitation example in Fig. 11(b) is more difficult because the bulk does not only induce a faulty line but shrinks the length property of the horizon as well. This issue occurs on all algorithms we tested. In our case, however, such an issue is mitigated thanks to the OHM; in Fig. 11(c), the positional error gap $Y^{\epsilon}$ between the true horizon (in yellow) and the faulty line (in pink) induced by ship edges is larger than the other two errors. This gap allowed the OHM to detect and replace the incorrect horizon line with the correct one.

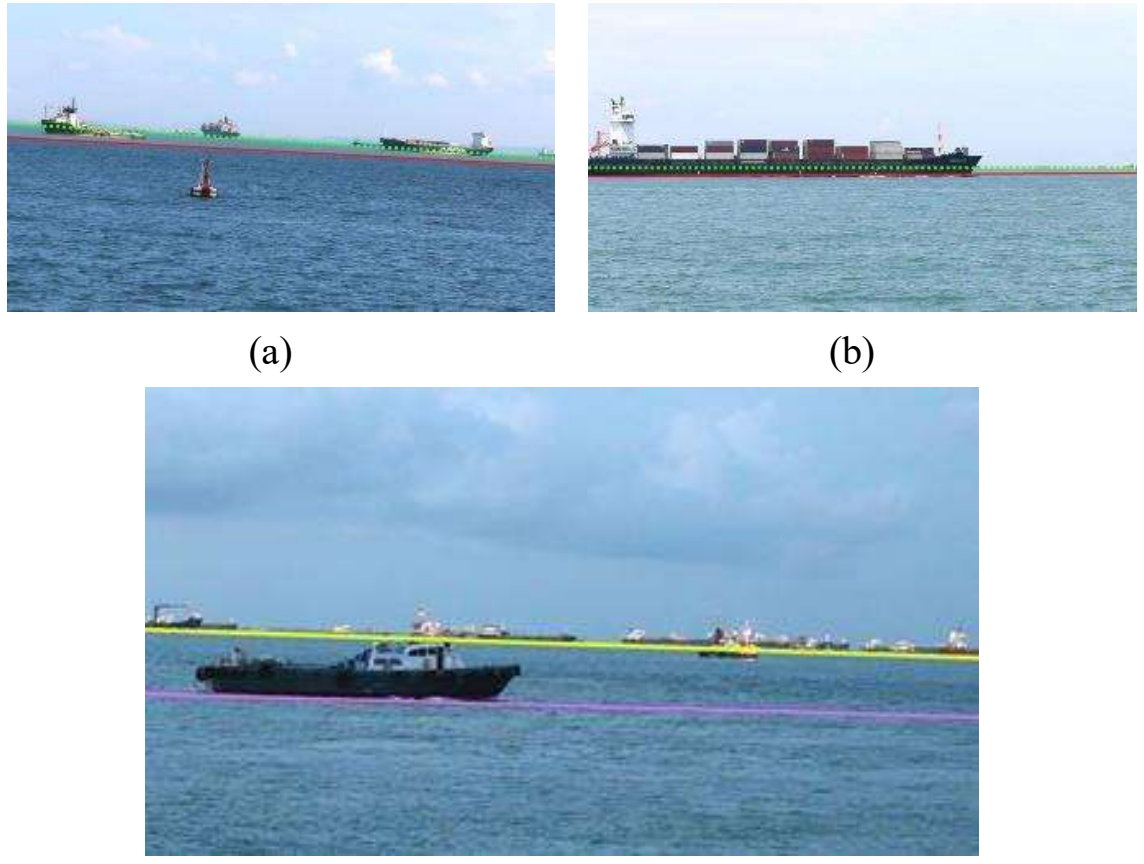

(a) (b) (c)

Fig. 11. Limitation examples: in (a) and (b), the green dotted line is the ground truth horizon, and the red line is the incorrect detection of our method; in (c), the incorrect detection (in pink) is successfully replaced with the correct line (in yellow) thanks to the OHM.

## V. CONCLUSIONS AND FUTURE WORK

The sea horizon is a critical element in maritime video processing, and detecting it under a variety of conditions, especially when the horizon edge response is weak, is a challenging task. Our method's key improvement is its filtering approach, which preserves weak horizon edges while exploiting the properties of line segments detected with a low edge threshold. We also addressed the computational constraint by proposing a vectorized implementation for efficient CPU execution and by accurately projecting the detected horizon from the downsized image to the original size. In addition, we have made a new maritime dataset (the TMD) publicly available, which is enriched with images from various sea conditions. Our algorithm's modular design will facilitate the integration of its functional stages, such as our robust outlier handler module, into other researchers' work. Our method is limited by noisy lines with stable gradient orientation or heavily occluded horizons. We have partially mitigated this limitation by incorporating temporal information, and we believe that further enhancements are attainable. Specifically, our algorithmic pipeline is readily scalable to integrate a Convolutional Neural Network (CNN)-based classification stage, utilizing the edges output by our filter as input and subsequently categorizing them as either horizon or noise. We think that the fusion of our filter's speed and robustness with a CNN will not only address the algorithm's limitations but also maintain an optimal execution speed, which we justify by the filter significantly constraining the number of input edges requiring classification by the CNN.

## CONFLICT OF INTEREST

The authors declare no conflict of interest.

## AUTHORS CONTRIBUTIONS

Y. Z. conducted the literature survey, proposed and developed the algorithmic solution, suggested and annotated the Tangier Maritime Dataset, designed the experimental setup, implemented the proposed and state-of-the-art algorithms, and wrote the paper. M. B guided the different research stages, contributed to the literature survey, directed the paper topic, and thoroughly reviewed the submitted manuscript. M. M guided the different research stages and thoroughly reviewed the submitted manuscript. A. A. proposed the research and paper topic, initiated the filter idea, guided the different research stages, reanalyzed and reviewed the results and interpretations, and thoroughly reviewed the submitted manuscript. The aforementioned author contributions are not exhaustive. All authors verified and approved the final version of the manuscript.

## Data Availability

Source code and dataset link will be available on GitHub: https://github.com/Zardoua-Yassir/A_fast_horizon_detector_and_a_new_annotated_dataset_for_maritime_video_processing.git

## Ackowledgement

We would like to express our gratitude to Sailor, Mr. Mustapha Tayebioui, for his invaluable assistance in collecting the Tangier Maritime Dataset.